\documentclass[final,5p,times,twocolumn,authoryear]{elsarticle}

\usepackage{amssymb}
\usepackage{amsmath}
\usepackage{multirow} 
\usepackage{bm}
\usepackage{times}
\usepackage{soul}
\usepackage{url}
\usepackage[hidelinks]{hyperref}
\usepackage[utf8]{inputenc}
\usepackage[small]{caption}
\usepackage{subcaption}
\usepackage{graphicx}
\usepackage{amsthm}
\usepackage{booktabs}
\usepackage{algorithm}
\usepackage{algorithmic}
\usepackage{xcolor} 
\usepackage{makecell}
\usepackage[switch]{lineno}
\usepackage{pifont}
\newcommand{\cmark}{\ding{51}}%
\newcommand{\xmark}{\ding{55}}%
\DeclareMathOperator*{\argmin}{argmin}

\journal{}

\begin{document}

\begin{frontmatter}



\title{Cross-Modal Prototype based Multimodal Federated Learning under Severely Missing Modality} 



\cortext[cor1]{Corresponding author.}
\author[label1]{Huy Q. Le }
\ead{quanghuy69@khu.ac.kr}
\author[label1]{Chu Myaet Thwal }
\ead{chumyaet@khu.ac.kr}
\author[label2]{Yu Qiao}
\ead{qiaoyu@khu.ac.kr}
\author[label1]{Ye Lin Tun}
\ead{yelintun@khu.ac.kr}
\author[label3]{ Minh N. H. Nguyen }
\ead{nhnminh@vku.udn.vn}
\author[label1]{Eui-Nam Huh}
\ead{johnhuh@khu.ac.kr}
\author[label1]{Choong Seon Hong \corref{cor1}}
\ead{cshong@khu.ac.kr}

\affiliation[label1]{organization={Department of Computer Science and Engineering, Kyung Hee University},
            city={Yongin-si},
            postcode={17104}, 
            country={Republic of Korea}}
\affiliation[label2]{organization={Department of Artificial Intelligence, Kyung Hee University},
            city={Yongin-si},
            postcode={17104}, 
            country={Republic of Korea}}
\affiliation[label3]{organization={Digital Science and Technology Institute, The University of Danang—Vietnam-Korea University of Information and Communication Technology},
            city={Da Nang},
            postcode={550000}, 
            country={Vietnam}}

\begin{abstract}
Multimodal federated learning (MFL) has emerged as a decentralized machine learning paradigm, allowing multiple clients with different modalities to collaborate on training a global model across diverse data sources without sharing their private data. However, challenges, such as data heterogeneity and severely missing modalities, pose crucial hindrances to the robustness of MFL, significantly impacting the performance of global model. The occurrence of missing modalities in real-world applications, such as autonomous driving, often arises from factors like sensor failures, leading knowledge gaps during the training process. Specifically, the absence of a modality introduces misalignment during the local training phase, stemming from zero-filling in the case of clients with missing modalities. Consequently, achieving robust generalization in global model becomes imperative, especially when dealing with clients that have incomplete data. In this paper, we propose \textbf{Multimodal Federated Cross Prototype Learning (MFCPL}), a novel approach for MFL under severely missing modalities. Our MFCPL leverages the complete prototypes to provide diverse modality knowledge in modality-shared level with the cross-modal regularization and modality-specific level with cross-modal contrastive mechanism. Additionally, our approach introduces the cross-modal alignment to provide regularization for modality-specific features, thereby enhancing the overall performance, particularly in scenarios involving severely missing modalities. Through extensive experiments on three multimodal datasets, we demonstrate the effectiveness of MFCPL in mitigating the challenges of data heterogeneity and severely missing modalities while improving the overall performance and robustness of MFL.
\end{abstract}


\end{frontmatter}




\section{Introduction}

\begin{figure}[]
	\centering
	\includegraphics[width=0.9\linewidth]{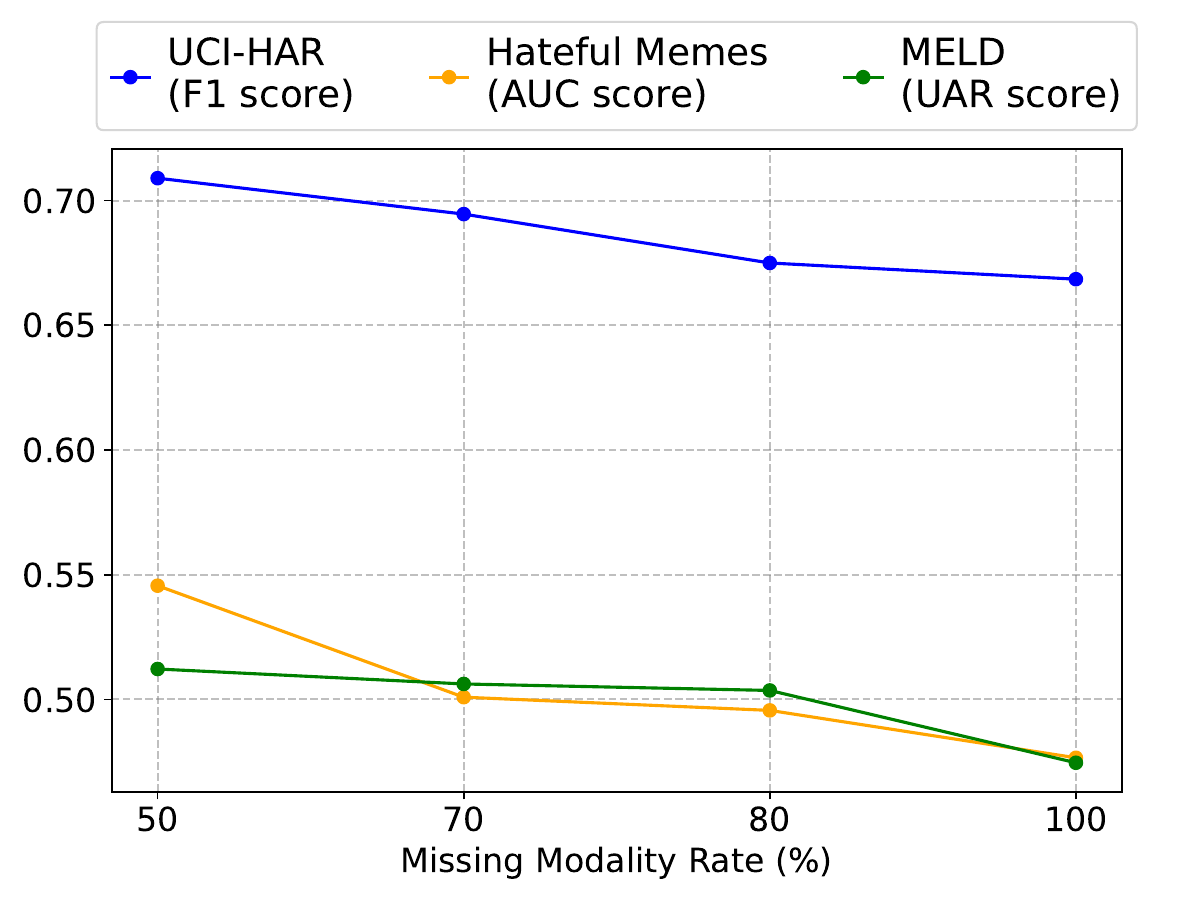}
	\caption{\textcolor{black}{\textbf{{Effect of different missing modality rates $q$} on FedAvg on different datasets.}}}
	\label{effect_missing} 
\end{figure}

Federated learning (FL)~\citep{mcmahan2017communication,kairouz2021advances,tan2022towards} has emerged as a promising distributed machine learning paradigm, allowing multiple clients to collaboratively train models without leaking private data. Despite promising outcomes observed across various domains~\citep{rieke2020future,nguyen2021federated}, most FL methods are tailored for scenarios with clients processing unimodal data, potentially constraining their applicability in real-world settings. However, in the era of data proliferation, users have the distinctive opportunity to harness many data types, unlocking new dimensions to explore the potential of multimodal data that aligns more closely with real-world applications~\citep{caesar2020nuscenes,fu2023tell}. Therefore, with the necessity of handling multimodal data while preserving privacy, multimodal federated learning (MFL) has emerged as a promising paradigm~\citep{xiong2022unified,zhao2022multimodal}. 
In traditional FL scenarios, data heterogeneity~\citep{li2020federated,deng2020adaptive}, where distributed data is non-IID (identically and independently distributed), has become a crucial challenge, leading to performance degradation. Despite numerous studies in FL domain~\citep{luo2021no,mendieta2022local} aimed at mitigating data heterogeneity, their predominant focus has been on addressing this challenge in applications involving unimodal data, rather than extending these solutions to the context of MFL. Additionally, in practical applications, it is essential to recognize that real-world scenarios may not always allow the utilization of all available modalities, particularly in MFL, where the access to complete modalities by clients is often not guaranteed. For example, if activities are recorded using two sensors and one experiences a disconnection (e.g., Bluetooth connectivity issues in human activity recognition or malfunctioning camera sensors in autonomous driving), it leads to a missing modality. Considering another scenario in text and image classification, some clients may only have access to image data (e.g., image without the caption), while others may possess both image and accompanying image data (e.g., images uploaded along with a caption). Such missing modalities could lead to a significant performance degradation, as the model trained under incomplete modality condition fails to capture the knowledge of multimodal features. \textcolor{black}{For example, on different multimodal datasets, FedAvg experiences a significant decrease in performance when the missing rate increases from $q=0.5$ to $q=1.0$, as shown in Figure~\ref{effect_missing}.} Challenges such as sensor disconnections, different hardware platforms, or other operational issues can impede the efficient integration of diverse data streams, impacting the model's robustness.
\begin{figure}[]
	\centering
	\includegraphics[width=0.9\linewidth]{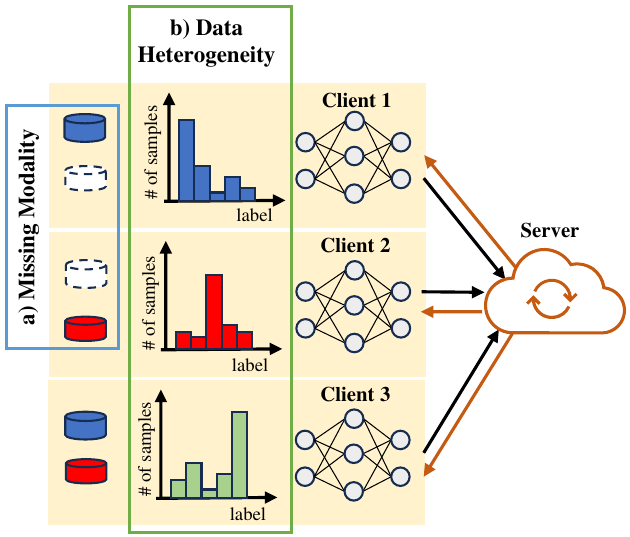}
	\caption{\textbf{Problem Illustration of Multimodal Federated Learning with $M=2$.} We present two challenges of MFL which are a) Missing Modality and b) Data Heterogeneity. }
	\label{problem} 
\end{figure}

While data heterogeneity presents its own set of challenges, the situation is compounded by the significant impact of severely missing modalities, leading to a further degradation in the model's performance within the context of MFL~\citep{chen2022fedmsplit,feng2023fedmultimodal}. Consequently, we pose a research question: \textit{How can we effectively address both data heterogeneity and missing modality challenges in MFL?}. In Figure~\ref{problem}, we illustrate the concepts of data heterogeneity and missing modalities. \textcolor{black}{Following FedMultimodal~\citep{feng2023fedmultimodal}, we set the equal missing rate for each modality on each client, creating a scenario of missing modalities. \textcolor{black}{We also experiment with scenarios where clients have partial access to data from missing modalities. Instead of assuming that missing modalities are completely unavailable, we consider cases where clients\footnote{\textcolor{black}{Individual entities (e.g, devices or nodes.) in federated learning system that own local data}} can access a specified proportion of the missing data, such as 20\% or 50\%. For example, in text and image classification task, clients may gain access to a designated portion of missing text data or image data. Similarly, when multiple accelerometer sensors are deployed, and some are disconnected in human activity recognition, clients can still utilize the data from the operational sensors.} We will further this experimental setting in subsection~\ref{performance_comparison}.}

\textcolor{black}{When dealing with significant missing modalities, a common practice is to impute the missing data with zeros. However, this zero-filling strategy introduces potential misalignments or undesired effects, adversely impacting the model performance in multimodal learning.} To enhance representation learning in MFL under conditions of missing modalities, an effective guiding method during training is crucial. Recently, prototype learning has been introduced as an effective information carrier between the server and clients in FL~\citep{tan2022federated,qiao2023mp}. A prototype is defined as the mean value of features from the same class, represented as a vector. \textcolor{black}{Prototypes are constructed by grouping feature vectors based on class labels and calculating their mean within a shared embedding space. This results in a compact and representative feature for each class in a low-dimensional feature space. Prototypes naturally reduce noise and outliers by averaging the feature vectors within a class, leading to robust and stable representations. Furthermore, they serve as lightweight class representatives by summarizing the meaningful core features of a class into a single compact vector, significantly reducing data size while retaining the most salient information
~\citep{snell2017prototypical,tan2022fedproto,zhang2024fedtgp}.} Several studies in FL have leveraged prototypes to tackle issues such as data heterogeneity~\citep{tan2022fedproto} or domain shift~\citep{huang2023rethinking}. Nonetheless, prior research on FL with prototypes has solely focused on scenarios involving unimodal data. Building upon the inspiration derived from prototype learning, we aim to extend its application to address the challenge of missing modalities in MFL. To tackle the challenges of both data heterogeneity and the existence of missing modalities, we derive the complete prototype by aggregating local prototypes from various clients, including those with both missing and full modalities, utilizing the shared projection head. This ensures that the features of each class encapsulate knowledge from diverse modalities. Additionally, the complete prototype serves as guidance for clients experiencing missing modalities.

In this paper, we propose \textbf{Multimodal Federated Cross Prototype Learning (\textbf{MFCPL})}, which consists of three main components: \textit{Cross-Modal Prototypes Regularization (CMPR), Cross-Modal Prototypes Contrastive (CMPC)} and \textit{Cross-Modal Alignment (CMA)}. Drawing inspiration from the success of prototypes in addressing data heterogeneity in FL~\citep{tan2022fedproto,mu2023fedproc}, we introduce a novel complete prototype to effectively handle missing modalities in MFL. First, we propose Cross-Modal Prototypes Regularization (CMPR) by  shortening the distance between complete prototype and the local modality-shared representation, bridging the gap between local and global models. Through this process, diverse modality knowledge is effectively transferred from complete prototypes to those clients with missing modalities, thereby contributing to the robustness of the learning system under conditions of missing modalities. Second, we introduce Cross-Modal Prototypes Contrastive (CMPC), leveraging complete prototypes to facilitate cross-modal contrastive learning. CMPC encourages local modality-specific representations to be close to the complete prototypes within the same semantic class and pushes them away from complete prototypes of different classes. Cross-modal contrastive learning involves aligning modality-specific representations and complete prototypes into the same latent space through a projection layer. We can provide robust and rich semantic representations suitable for clients with missing modalities by aligning the complete prototype with modality-specific representations. Finally, we introduce Cross-Modal Alignment (CMA), which aligns the embedding features of the missing modality with those of the existing modality through a projection layer. By incorporating this, we mitigate the negative impact of missing modality and improve the generalization performance of the global model. These components collectively empower our proposed framework as a robust method to tackle severe missing modalities in heterogeneous MFL effectively. The details of each component are provided in subsection~\ref{MFCPLsection}. Our primary contributions are:
\begin{itemize}
    \item To the best of our knowledge, we are the first in heterogeneous MFL to integrate the concept of complete prototype for addressing the missing modality challenge. Leveraging the prototype perspective, we enhance performance by calibrating the local training process with diverse modality knowledge, particularly benefiting clients dealing with missing data.
    \item We propose a novel multimodal federated learning framework, \textbf{MFCPL}, to learn the global model with severe missing modality. Building on the success of prototypes in FL, we introduce complete prototypes aligned at two levels, offering rich semantic information from multiple modalities: modality-specific representation with CMPC and modality-shared representation with CMPR through a projection layer. Additionally, we address the negative impact associated with missing modality by introducing the loss function for CMA.
    \item We comprehensively evaluate the performance of MFCPL through extensive experiments on three multimodal federated datasets, demonstrating its superiority over other baseline methods. Complemented by ablation studies, the promising results validate the effectiveness of each component.
\end{itemize}
\begin{figure*}[]
	\centering
    \includegraphics[width=\linewidth]{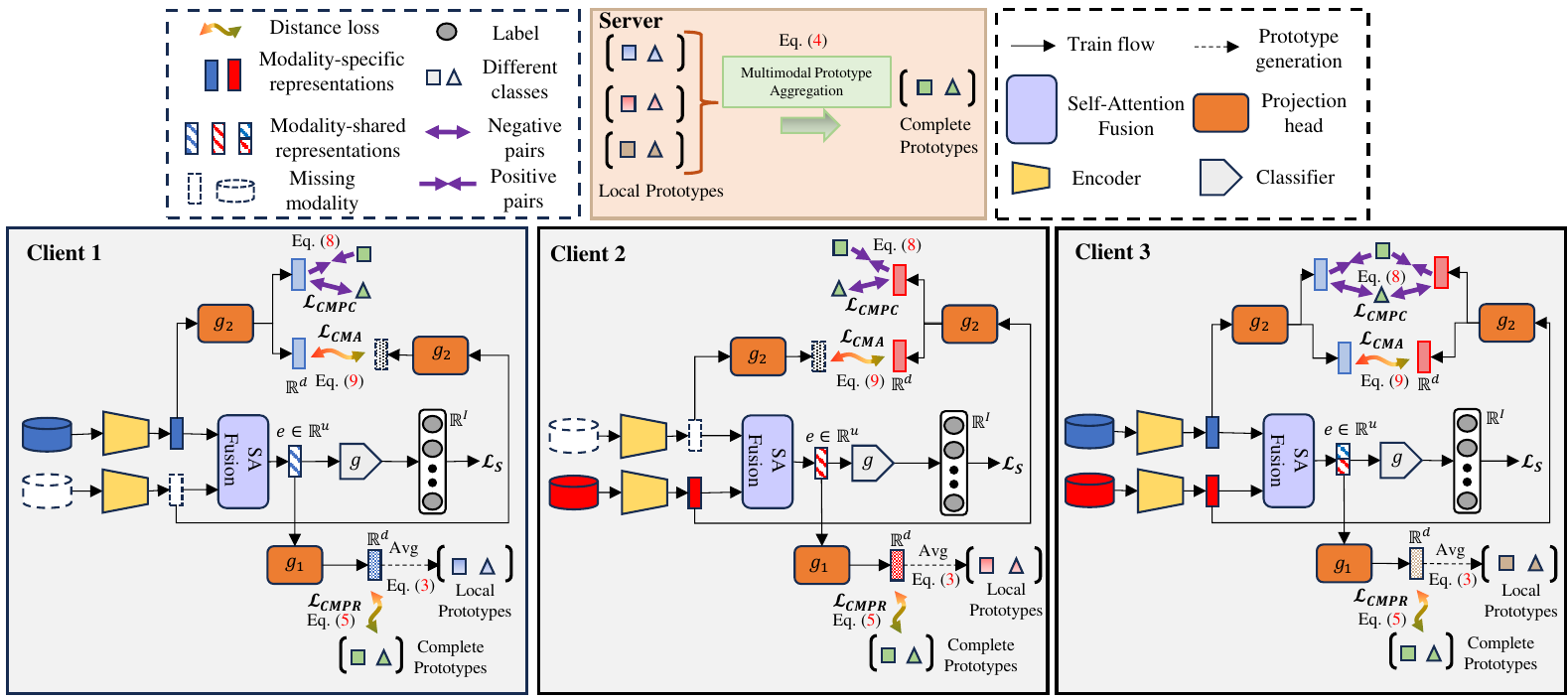}
	\caption{\textcolor{black}{\textbf{Illustration of \textbf{MFCPL} with $M=2$.} Different client types upload their local prototypes based on Eq.~\ref{proto_1} to server using projection head $g_1$. We provide diverse modality knowledge with the complete prototypes from Eq.~\ref{aggproto} in modality-shared level with $\mathcal{L}_{CMPR}$ from Eq.~\ref{eq:cmpr} and modality-specific level with $\mathcal{L}_{CMPC}$ from Eq.~\ref{cmpc} using projection head $g_2$. The alignment loss $\mathcal{L}_{CMA}$ in Eq.~\ref{cma} is introduced to reduce the negative effect arising from zero-filling of missing modalities and enhance the coherence between projected modality-specific representations of two modalities using projection head $g_2$. In the illustration, the square class represents the positive class, while the triangle class represents the negative class.}}
	\label{system} 
\end{figure*}

\section{Related Work}
\subsection{Multimodal Federated Learning}
Numerous research works in multimodal federated learning (MFL)~\citep{xiong2022unified,zhao2022multimodal,yu2023multimodal}, have explored diverse techniques to facilitate the decentralized training of machine learning models across multiple data sources. For instance,~\citep{xiong2022unified} introduced a co-attention layer to merge representations from various modalities, obtaining global features to train personalized models for clients. \textcolor{black}{In a different approach, the authors in~\citep{yang2022cross,yang2024cross} addressed the task of cross-modal federated human activity recognition (HAR) by proposing a feature disentangled activity recognition network to capture modality-discriminative characteristics for each client. However, their work focuses on a cross-modal setting with HAR task, where each client possesses data from only one modality. In contrast, multimodal federated learning considers scenarios where clients may have access to multiple modalities. In our work, by adjusting the missing rate, we extend our approach to accommodate both cross-modal and multimodal setting.} Addressing labeled data constraints in MFL,~\citep{zhao2022multimodal} proposed a semi-supervised framework within the FedAvg mechanism~\citep{mcmahan2017communication}. CreamFL~\citep{yu2023multimodal}, FedMEKT~\citep{le2024fedmekt} introduced a knowledge distillation-based MFL framework for representation aggregation leveraging the public dataset. However, these above method rely on additional public datasets, which is impractical in real-world scenarios. Moreover, these methods did not consider the missing modality scenario in the system, limiting their practical applicability in real-world settings. To address the issue of missing modalities in MFL,~\citep{chen2022fedmsplit} employed a dynamic graph structure to capture correlations among different types of clients. Recently, FedMultimodal~\citep{feng2023fedmultimodal} introduced the first benchmark in MFL that evaluates MFL under different types of data corruption: missing modalities, missing labels, and erroneous labels. In this paper, to address the missing modality problem in heterogeneous MFL, we introduce the novel concept of leveraging a complete prototype to operate with local representations at two levels: \textit{modality-specific representations} and \textit{modality-shared representation} through a projection layer.

\subsection{Incomplete Multimodal Learning}
Exploring incomplete data, such as missing modalities in multimodal learning, is crucial while ensuring robustness in real-world scenarios with missing modalities. Instances of missing modalities are prevalent in practical situations, often stemming from hardware constraints or storage limitations. \textcolor{black}{SMIL~\citep{ma2021smil} proposed a Bayesian meta-learning framework that incorporates the missing modality reconstruction and feature regularization to tackle the problem of multimodal learning with severely missing modalities.} Moreover, various approaches, such as data imputation~\citep{tran2017missing} or data reconstruction~\citep{pham2019found,zhao2021missing}, have been proposed to address these challenges by filling in missing modalities based on existing ones. \textcolor{black}{For instance,~\citep{tran2017missing} proposed a cascaded residual autoencoder (CRA), which utilized stacked residual autoencoders to learn the complex relationships between different modalities to impute missing ones.} Regarding model robustness, the authors in~\citep{ma2022multimodal} investigate the Transformer robustness against modal-incomplete data and improve it via multi-task optimization. Additionally,~\citep{dong2023simmmdg} introduced a cross-modal translation method to replace null entries in the embedding for missing modality generalization during the inference stage. Another approach involves utilizing a knowledge transfer mechanism to convey information from modalities with available data to those with missing data~\citep{garcia2018modality,luo2018graph}.~\citep{poklukar2022geometric} introduced the Geometric Multimodal Contrastive (GMC) framework, which geometrically aligns complete and modality-specific representations to address missing modality information across various tasks. In this study, we employ the definition of a complete prototype to provide diverse modality knowledge for clients with missing modalities, thereby enhancing the local training process for these clients.

\subsection{Prototype Learning}
The prototype is computed as the mean feature vectors within the same class~\citep{snell2017prototypical} and has found applications in various research domains, including few-shot learning~\citep{xu2020attribute,zhang2021prototype}, and unsupervised learning~\citep{ye2019unsupervised,guo2022hcsc}. \textcolor{black}{Prototypes provide robust class representations by reducing noise and mitigating the influence of outliers.} In FL, prototypes have been employed in multiple studies to address data heterogeneity~\citep{tan2022fedproto,tan2022federated,huang2023rethinking,qiao2023mp}. FedProto~\citep{tan2022fedproto} was among the pioneering works to apply prototypes in FL, introducing a communication approach between clients and the server using prototypes instead of parameter exchange to learn personalized models. Additionally, ~\citep{qiao2023mp} proposed the multi-prototype approach with the contrastive learning strategy to address the feature heterogeneity. Recently, FPL~\citep{huang2023rethinking} proposed the cluster prototypes to tackle the domain shift in FL. However, the majority of prior works concentrate on scenarios where clients exclusively possess unimodal data. In the context of MFL under missing modalities, it is crucial to address the generalization across various types of clients with diverse modality absences. Our work illuminates the use of the complete prototype to accomplish this objective at two levels: modality-specific and modality-shared representations.

\section{Methodology}
\subsection{Preliminaries}
Multimodal federated learning (MFL) aims to facilitate collaborative training among multiple clients on a shared global model, leveraging multimodal data while ensuring data privacy. \textcolor{black}{In this paper, we consider that there are $N$ clients (indexed by $i$) with private multimodal data $D_i=\{{x}^1_i,{x}^2_i,...,x^M_i,{y}_i\}$, where $x^m_i$ represents samples from the modality $m \in M$ and $y_i$ denotes the label associated with each sample. In scenarios with missing modalities, the combination of $M$ modalities leads to $2^M-1$ possible types of clients, as each client may possess data corresponding to a unique subset of modalities. For example, for applications with image and text modalities, there are three types of client: clients missing image data, clients missing text, and clients with both image and text data.}

Client models share the identical architecture, comprising three modules: specific encoder of each modality  $f_m$, self-attention fusion module $h_{\text{self-attention}}$, and classifier $g$. \textcolor{black}{Each encoder $f_m$ processes input samples from modality $m$ to generate \(n\)-dimensional feature vectors \(z_m = f_m(x^m)\in\mathbb{R}^n\). Subsequently, these modality-specific representations are then integrated into a unified multimodal representation by the self-attention fusion module $h_{\text{self-attention}}$, denoted as $e=h_\text{self-attention}([z_1,z_2,...z_M])\in \mathbb{R}^u$. Here, $[z_1,z_2,...z_M]$ represents the concatenation of modality-specific representations.} Finally, the classifier $g$ maps the multimodal feature $e$ to the logits output $z_{cls}=g(e)\in \mathbb{R}^I$. Specifically, given the model parameters for the whole network as $\theta$, and $D$ is the sum of samples over all clients, we formulate the global objective:
\begin{align}
    \argmin_{\theta}{L}(\theta) &= \sum^N_{i=1}\frac{|D_{i}|}{|D|}{\mathcal{L}}_{i}(\theta_{i},D_i),
\end{align}
where the loss function $\mathcal{L}_i$ is the task training loss $\mathcal{L_S}=L(z_{cls},y)$ for $i^{th}$ client.
\subsection{Multimodal Prototype Federated Learning}
Previous research work on federated prototype learning~\citep{tan2022fedproto,mu2023fedproc} has focused on scenarios in which clients possess unimodal data. \textcolor{black}{In the context of MFL with $M$ modalities and missing modality scenarios, multiple types of clients arise, such as those missing one or more modalities and those with $M$ modalities available.} Simply aggregating unimodal prototypes based on modality may introduce a bias toward the dominant modality and cause the performance degradation in FL under severely missing modalities. This prompts us to leverage prototypes from all modalities to develop a generalized model for different types of clients. 
\textcolor{black}{\paragraph{Complete Prototypes}To facilitate alignment in the local training process, we utilize the projection head $g_1$ to map the multimodal representations $e$ from different client types into the \textit{shared representation space $\mathbb{R}^d$}: 
\begin{align}
\{r_{\Omega_i}=g_1(e_{\Omega_i})\}\in\mathbb{R}^d. 
\end{align}
where ${\Omega_i} \subseteq \Omega=\{1,2,...,M\}$ denotes the subset of modalities available to a given client type. We define the $k^{th}$ class prototype from different types of clients as:
\begin{align}
    p^k_{{\Omega_i}}=\frac{1}{|S^k_{{\Omega_i}}|}\sum_{i\in S^k_{{\Omega_i}}}r_{{\Omega_i}},
\label{proto_1}
\end{align}
where $S^k_{{\Omega_i}}$ represents the subset of $D_i$ belonging to $k^{th}$ class of different client types of $\Omega_i$. As complementary to unimodal prototypes, we introduce complete prototypes by averaging local prototypes from different types of clients in the shared representation space $\mathbb{R}^d$. Specifically, we formulate the complete prototypes on the server as follows:
\begin{align}
    \begin{split}
    P^k&=\frac{1}{N}\sum_{i\in N}\{p^k_{{\Omega_i}} |\Omega_i \subseteq \Omega\}\in\mathbb{R}^d 
    \\
    P&=[P^1,P^2,\dots,P^K],
    \end{split}
    \label{aggproto}
\end{align}
where $P^k$ denotes the complete prototypes belonging to class $k\in K$. Specifically, our complete prototypes guide the local training process with the calibration at two levels: modality-specific and modality-shared representations.}

\begin{algorithm}[t]
    {\color{black}
    \caption{\textbf{MFCPL}}
    \label{alg:algorithm}
    \textbf{Input}: communication rounds T, local epochs R, number of local clients N, local dataset $D_i$.\\
    \textbf{Output}: Final global model $\theta_t$
    \begin{algorithmic}[1] 
    \STATE \textbf{Server Execution:} 
    \FOR{$t=0,\dots,T$}
    \FOR{$i=0,\dots,N$} 
    \item  $\theta^i_t, p_{{\Omega_i}}\leftarrow\textbf{LocalUpdate}(\theta_t,P)$
    \ENDFOR

    $ P^k\leftarrow\frac{1}{N}\sum_{i\in N}\{p^k_{{\Omega_i}} |\Omega_i \subseteq \Omega\}$ by Eq.~(\ref{aggproto}) \\
    \vspace{0.15cm} 
    $\theta_{t+1}\leftarrow\sum_{i=1}^N\frac{|D_i|}{|D|}\theta_t^i$
    \ENDFOR
    \STATE \textbf{Client Execution:}
    \STATE \textbf{LocalUpdate}($\theta_t,P$):
    \FOR{$r=0,\dots,R$}
    \FOR{each batch $\in D_i=\{D_i=\{{x}^1_i,{x}^2_i,...,x^M_i,{y}_i\}$}
    \STATE $z_m = f_m(x^m)$
    \STATE $e=h_\text{self-attention}([z_1,z_2,...z_M]),r_{\Omega_i}=g_1(e_{\Omega_i})$
    \STATE $z'_{m_i}=g_2(z_{m_i})$
    \STATE $\mathcal{L}_{CMPR}\leftarrow(r_{\Omega_i},P)$ in Eq.~(\ref{eq:cmpr})
    \STATE $\mathcal{L}_{CMPC}\leftarrow(z'_{m_i},P)$ in Eq.~(\ref{cmpc})
    \STATE $\mathcal{L}_{CMA}\leftarrow(z'_{m_i}, \forall m\in \{1,2,...,M\} )$ in Eq.~(\ref{cma})
    \STATE $\mathcal{L}_{S}\leftarrow(z_{cls}^i,y_i)$
    \STATE $\theta^i_t\leftarrow \theta^i_t-\eta\nabla\mathcal{L}$
    \ENDFOR
    \ENDFOR
    \STATE $p_{\Omega_i}=[p^1_{\Omega_i},p^2_{\Omega_i},\dots,p^K
    _{\Omega_i}]$
    \STATE \textbf{return} $\theta^i_t,p_{\Omega_i}$      
    \end{algorithmic}
    
\label{mainalg}
}
\end{algorithm}

\subsection{Multimodal Federated Cross Prototype Learning}~\label{MFCPLsection}
To achieve a generalized model for different types of clients under the condition of missing modalities, we utilize complete prototypes to provide diverse modality knowledge, supporting local clients at two levels: modality-specific and modality-shared representations. As shown in Figure~\ref{system}, our proposed method comprises three main components: Cross-Modal Prototypes Regularization (CMPR), Cross-Modal Prototypes Contrastive (CMPC), and Cross-Modal Alignment (CMA).
\paragraph{Cross-Modal Prototypes Regularization} Similar to other traditional FL methods, MFL encounters the challenge of data heterogeneity, and further complicated by the issue of missing modalities. In response to these challenges, we introduce a regularization loss that narrows the gap between local and global models, simultaneously providing diverse modality knowledge for clients with missing modalities at the modality-shared representation level. \textcolor{black}{Drawing inspiration from FedProto~\citep{tan2022fedproto}, we incorporate a regularization loss between local projected multimodal representations $r_{\Omega_i}$ and complete prototypes $P^k$ within the same semantic class $k$}. However, our approach differs from FedProto, as our regularization loss serves as an extension to address the challenges posed by missing modalities in the context of multimodal federated learning. To implement this regularization, we utilize $\ell_2$ distance and introduce the Cross-Modal Prototypes Regularization (CMPR) as follows:
\textcolor{black}{
\begin{align}
    \mathcal{L}_{CMPR}=\sum_{k}\|r^k_{\Omega_i}-P^k\|_2^2,
    \label{eq:cmpr}
\end{align}}
\textcolor{black}{where $r^k_{\Omega_i}$ is the local projected multimodal representations of $k$ semantic class.} \textcolor{black}{By conducting calibration on the modality-shared representation with complete prototypes, we alleviate the impact of missing modalities from the multimodal perspective. \textcolor{black}{Additionally, the regularization reduces the divergence between local models and global model, thereby mitigating data heterogeneity.} Furthermore, we maintain the coherence between the modality-shared representation and complete prototypes in the shared space through the projection head.}   

\paragraph{Cross-Modal Prototypes Contrastive} We envision that modality-specific representations in clients with missing modalities can capture rich and diverse modality knowledge from the complete prototype. The modality-specific features of local data instances should align closely with their respective complete prototypes within the same semantic class while exhibiting dissimilarity to complete prototypes associated with different semantic classes. \textcolor{black}{For data instances $D_i=\{{x}^1_i,{x}^2_i,...,x^M_i,{y}_i\}$, we encode each modality $m$ using its respective encoder $f_m$ to acquire the modality-specific feature vectors as follows:
\begin{align}
    z_{m_i}=f_m({x}^m_i)\in\mathbb{R}^n, \forall m\in \{1,2,...,M\}. 
\end{align}
}
\textcolor{black}{To facilitate alignment between the modality-specific features and complete prototypes, we employ the projection head $g_2$ to project the modality-specific feature vectors of each modality into the shared representation space:
\begin{align}
z'_{m_i}=g_2(z_{m_i})\in\mathbb{R}^d, \forall m\in \{1,2,...,M\}.
\end{align}
}
\textcolor{black}{Let $p$ be the corresponding complete prototype $p\in P$, $C^k=P-P^k$ represents the prototypes with different semantic class from the local instances. Subsequently, the Cross-Modal Prototypes Contrastive (CMPR) loss where all modalities $M$ are available is expressed as follows:
\begin{align}
        \mathcal{L}_{CMPC} = \sum_{m=1}^{M}-\log\frac{\sum_{p\in P^k}\exp(s(z'_{m_i},p)/\tau)}{\sum_{p\in P}\exp(s(z'_{m_i},p)/\tau)}, 
          \label{cmpc}
\end{align}
where $\mathrm{s}( u, v) =  u^\top  v / \lVert u\rVert \lVert v\rVert$ denotes the cosine similarity between the projected modality-specific feature and complete prototypes, and $\tau$ is the contrastive hyper-parameter. In cases where clients lack data for modality $m$, the corresponding contrastive loss for that modality $m$ will be excluded.} The goal of  Eq.~(\ref{cmpc}) is to encourage the modality-specific representations from clients with missing modalities to closely align with the corresponding complete prototypes in the shared embedding space, thereby enhancing the generalization of these clients \textcolor{black}{and mitigate the negative impact of missing modality from the unimodal perspective.} \textcolor{black}{Through this calibration, we utilize the projection head to facilitate alignment between features of the complete prototype and modality-specific representations. Additionally, we ensure the preservation of relevance between modality-specific representations and the complete prototypes in the shared space.} The effectiveness of the projection head is further discussed in subsection~\ref{subsec:dimension}.
\begin{figure*}[t]
    \centering
    \begin{subfigure}{0.32\textwidth}
        \centering
        \includegraphics[width=\linewidth]{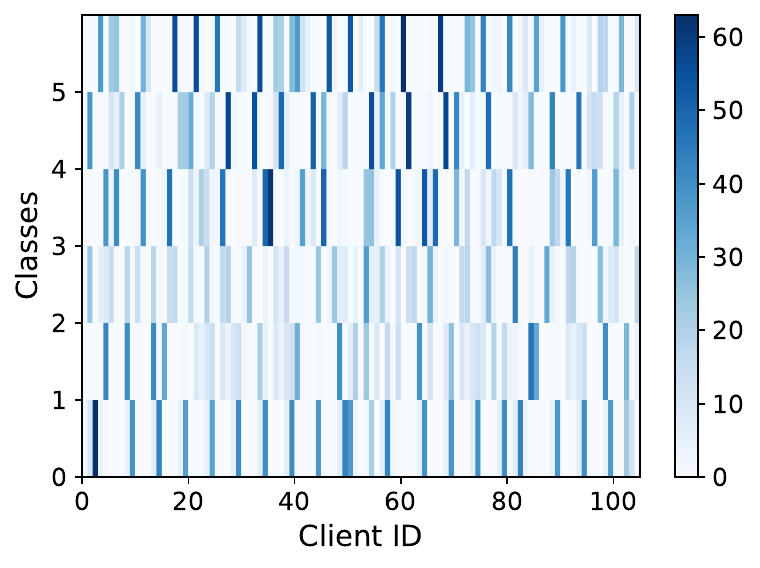}
        \caption{UCI-HAR}
        \label{fig:temperature_ucihar}
    \end{subfigure}%
    \hfill
    \begin{subfigure}{0.32\textwidth}
        \centering
        \includegraphics[width=\linewidth]{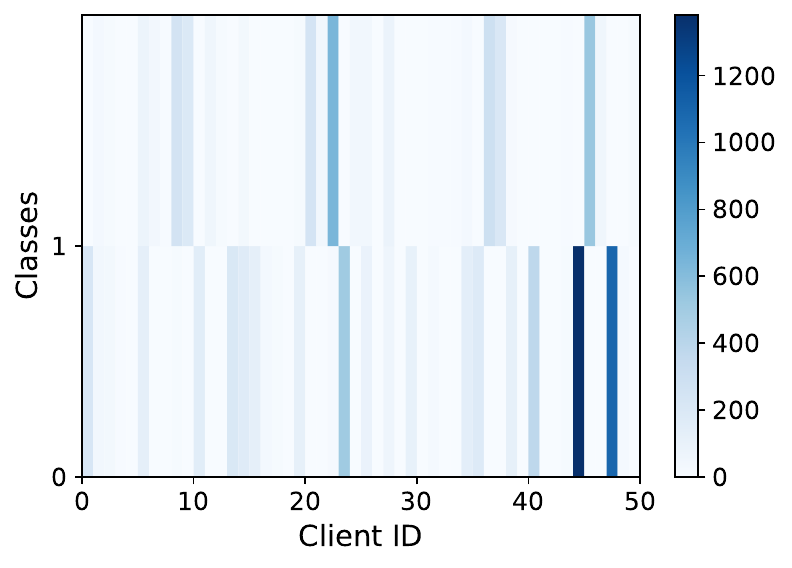}
        \caption{Hateful Memes}
        \label{fig:dimension_hatefulmemes}
    \end{subfigure}%
    \hfill
    \begin{subfigure}{0.32\textwidth}
        \centering
        \includegraphics[width=\linewidth]{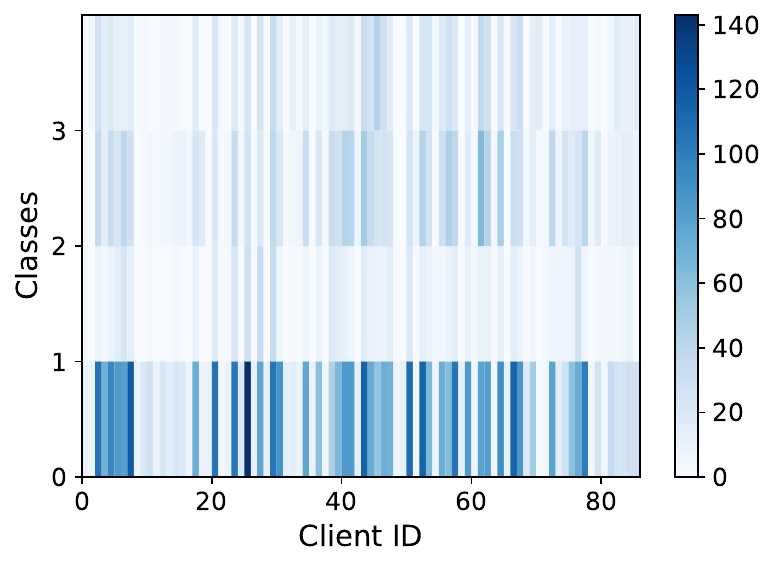}
        \caption{MELD}
        \label{fig:dimension_meld}
    \end{subfigure}
    \caption{\textbf{The data distribution of training data for each client with non-IID characteristic.} The color bar denotes number of data samples. Each rectangle defines the number of samples in each class of each client.}
    \label{fig:datadistribution}
\end{figure*}


\begin{table*}[ht]
\centering
\caption{\textbf{Client distribution} on UCI-HAR, Hateful Memes and MELD datasets under various missing rates. I,T,A,V represent Image, Text, Audio, Video modalities, respectively.}
\textcolor{black}{
\begin{tabular}{llp{12cm}}
\toprule
\textbf{Dataset}      & \textbf{Missing Rate} & \textbf{Client Distribution} \\ \midrule
\multirow{4}{*}{UCI-HAR}    
                      & 50\%                  & Acc: 35, Gyro: 40, Acc-Gyro: 30 \\ 
                      & 70\%                  & Acc: 43, Gyro: 55, Acc-Gyro: 7 \\ 
                      & 80\%                  & Acc: 54, Gyro: 48, Acc-Gyro: 3 \\ 
                      & 100\%                 & Acc: 55, Gyro: 50 \\ \midrule
\multirow{4}{*}{Hateful Memes} 
                      & 50\%                  & I: 20, T: 16, I-T: 14 \\ 
                      & 70\%                  & I: 21, T: 26, I-T: 3 \\ 
                      & 80\%                  & I: 30, T: 19, I-T: 1 \\ 
                      & 100\%                 & I: 29, T: 21 \\ \midrule
\multirow{4}{*}{MELD} 
                      & 50\%                  & A: 14, T: 13, V: 13, A-T: 14, A-V: 10, V-T: 10, A-T-V: 12 \\ 
                      & 70\%                  & A: 26, T: 22, V: 22, A-T: 4, 6: 12, V-T: 5, A-T-V: 1 \\ 
                      & 80\%                  & A: 28, T: 40, V: 23, A-T: 2, A-V: 3, V-T: 1 \\ 
                      & 100\%                 & A: 24, T: 32, V: 30 \\ \bottomrule
\end{tabular}
}
\label{tab:client_distribution}
\end{table*}

\paragraph{Cross-Modal Alignment}While the complete prototypes contribute diverse modality knowledge to address data heterogeneity and missing modalities at modality-specific and modality-shared representation levels, simply imputing missing data with zeros for the absent modality can introduce noise. We introduce an alignment loss between modality-specific representations from different modalities to alleviate negative effect arising from zero-filling of missing modalities. \textcolor{black}{By aligning features within the shared representation space, the CMA loss facilitates effective complementarity between features from different modalities.} We apply the alignment loss to the projected modality-specific representations. \textcolor{black}{To achieve this goal, we compute the alignment loss using the $\ell_2$ distance between the projected representations of all available modality pairs. For $M$ modalities, the Cross-Modal Alignment loss is formulated as:
\begin{align}
    \mathcal{L}_{CMA}=\sum_{(m_1,m_2)\in \mathcal{P}_M}\sum_{i\in D_i}\|z'_{m_{1i}}-z'_{m_{2i}}\|_2^2,
    \label{cma}
\end{align}
where $\mathcal{P}_M$ denotes the set of all pairs of modalities among available $M$ modalities, and $z'_{m_{1i}}=g_2(z_{m_{1i}})$, $z'_{m_{2i}}=g_2(z_{m_{2i}})\in\mathbb{R}^d$ represent the projected modality-specific representations for modalities $m_1$ and $m_2$, respectively.} Our objective is to align the representation from the absent modality with that of the existing modality, thereby mitigating the negative impact in the missing modality. \textcolor{black}{Moreover, since the original representation spaces of each modality is different from each other, alignment in the shared space enables the coherence across modality-specific representation. This approach has shown effectiveness in improving performance in multimodal learning~\citep{pham2019found,dong2023simmmdg}. Our experiments demonstrate that this alignment, as tailored in our design, significantly enhances the method's effectiveness and illustrates its adaptability across various loss functions.} Finally, the overall training objective for each client is formulated as follows:
\begin{align}
\mathcal{L}=\mathcal{L}_S+\alpha_{reg}\mathcal{L}_{CMPR}+\alpha_{con}\mathcal{L}_{CMPC}+\alpha_{align}\mathcal{L}_{CMA},
\label{overallloss}
\end{align}
where $\alpha_{reg}$, $\alpha_{con}$ and $\alpha_{align}$ are hyper-parameters that control the importance of the $\mathcal{L}_{CMPR}$, $\mathcal{L}_{CMPC}$ and $\mathcal{L}_{CMA}$ terms, respectively. The overall training process for MFCPL is shown in Algorithm~\ref{mainalg}. The server broadcasts the complete prototypes to all clients in each communication round\footnote{\textcolor{black}{Iterative steps in federated learning where clients update local models and send them to server for aggregation. Each round represents one cycle of this process.}}. In the local training phase, each client trains the learning model on private data using the overall training loss defined in Eq.~(\ref{overallloss}).

\section{Experiments}
\subsection{Experimental Setup}
\subsubsection{Datasets} In this paper, we perform expeirments on three multimodal federated datasets to evaluate the effectiveness of MFCPL in various tasks.
\begin{itemize}
    \item \textbf{UCI-HAR}~\citep{anguita2013public} dataset consists of smartphone sensory data from Accelerometer and Gyroscope sensors. This dataset involves observations from $30$ volunteers engaged in six distinct activities: walking, walking upstairs, walking downstairs, sitting, standing, and laying. 
    \item \textbf{Hateful Memes}~\citep{kiela2020hateful} dataset comprises $10,000$ samples of image and text pairs organized into binary classes. The primary objective is to identify and detect hateful speech within memes.
    \item \textbf{MELD}~\citep{poria2018meld} dataset is a multimodal conversational dataset for emotion recognition collected from Friends TV series, which contains about $13,000$ utterances from $1,433$ dialogues. \textcolor{black}{In this experiment, we utilize audio, visual and text modalities.} Following the setting from~\citep{feng2023fedmultimodal}, we keep $4$ emotion labels with most samples which are neutral, sadness, happiness and anger.
\end{itemize}
Following~\citep{feng2023fedmultimodal}, we initialize $50$, $86$, and $105$ clients with the participation rates of $0.25$, $0.1$, and $0.1$ per communication round for Hateful Memes, MELD, and UCI-HAR datasets, respectively. We illustrate the data partition of three datasets in Figure~\ref{fig:datadistribution}. To simulate the non-IID scenario, we apply the Dirichlet distribution~\citep{ferguson1973bayesian} with $\beta=0.2$ to the Hateful Memes and UCI-HAR datasets. In contrast, the MELD dataset inherently exhibits a natural non-IID data distribution due to its organization by speakerIDs. \textcolor{black}{We perform a random split for the UCI-HAR~\citep{anguita2013public} dataset with $70\%$ allocated for training data and $30\%$ for test data. Within the training set, $20\%$ is further set aside for the validation set. In the case of the Hateful Memes dataset, we adhere to the setup outlined in the original work of this dataset~\citep{kiela2020hateful}. Specifically, we construct a validation set and test set from $5\%$ and $10\%$ of total data, respectively, and use the remaining portion for training data. As for the MELD dataset, we adopt the train-test-validation split provided in the original work of this dataset~\citep{poria2018meld}. We utilize the validation procedure in model evaluation to select the optimal model and mitigate the risk of overfitting on the training set.}
\begin{table*}[t]
\caption{{\textbf{Performance comparison of our \textbf{MFCPL} with SOTA methods} on UCI-HAR, Hateful Memes and MELD datasets with various modality missing rates. The best results are marked in \textbf{bold}.}}
\centering
\resizebox{18.5cm}{!}{%
\begin{tabular}{c|cccc|cccc|cccc}
\hline
\multirow{2}{*}{Methods} & \multicolumn{4}{c|}{UCI-HAR (F1 score)}                     & \multicolumn{4}{c|}{Hateful Memes (AUC score)}               & \multicolumn{4}{c}{MELD (UAR score)}                         \\ 
                                  & $q = 0.5$        & $q = 0.7$        & $q = 0.8$        & $q = 1.0$        & $q = 0.5$        & $q = 0.7$        & $q = 0.8$        & $q = 1.0$        & $q = 0.5$        & $q = 0.7$        & $q = 0.8$        & $q = 1.0$        \\ \hline
FedAvg~\citep{mcmahan2017communication}                            & 70.90          & 69.46          & 67.50          &  \textcolor{black}{66.85}               & 54.56          & 50.09          & 49.56          &  \textcolor{black}{47.66}               &  \textcolor{black}{51.22}               & \textcolor{black}{50.62}                &  \textcolor{black}{50.36}               &  \textcolor{black}{47.46}               \\
FedProx~\citep{li2020federated}                           & 69.88          & 69.74          & 67.92          &  \textcolor{black}{67.23}               & 54.89          & 53.92          & 50.25          & \textcolor{black}{48.82}                &  \textcolor{black}{52.45}               &  \textcolor{black}{51.02}               &    \textcolor{black}{50.55}             &  \textcolor{black}{49.13}               \\
FedOpt~\citep{reddi2020adaptive}                            & 74.79          & 66.17          & 56.45          & \textcolor{black}{55.83}                & 54.62          & 51.61          & 49.58          &  \textcolor{black}{47.29}               &  \textcolor{black}{51.58}               & \textcolor{black}{49.27}                & \textcolor{black}{47.54}                & \textcolor{black}{46.61}                \\
MOON~\citep{li2021model}                              & 70.52          & 69.52          & 68.15          & \textcolor{black}{67.05}                & 53.45          & 52.02          & 51.75          & \textcolor{black}{49.36}                & \textcolor{black}{51.88}                & \textcolor{black}{51.10}                & \textcolor{black}{50.99}               & \textcolor{black}{49.32}                \\ \hline
FedProto~\citep{tan2022fedproto}                          & 74.87          & 73.63          & 72.78          & \textcolor{black}{70.99}                & 52.67          & 51.40          & 50.68          &  \textcolor{black}{47.15}               & \textcolor{black}{50.66}                & \textcolor{black}{47.51}                 &  \textcolor{black}{46.94}                & \textcolor{black}{36.80}                   \\
FedProc~\citep{mu2023fedproc}                      & 70.14          & 69.37          & 68.77          &   \textcolor{black}{65.43}              & 52.75          & 50.49          & 50.23          &  \textcolor{black}{48.81}               &  \textcolor{black}{51.92}                  & \textcolor{black}{51.83}                    & \textcolor{black}{50.48}                   &   \textcolor{black}{49.80}                 \\
FPL~\citep{huang2023rethinking}                             & 75.16          & 74.28          & 73.46          & \textcolor{black}{71.58}                & 54.97          & 53.79          & 52.94          &    \textcolor{black}{50.79}             & \textcolor{black}{52.11}                    & \textcolor{black}{50.09}                   & \textcolor{black}{49.49}                   & \textcolor{black}{48.76}                   \\ \hline
\textcolor{black}{FedIoT~\citep{zhao2022multimodal}}                           &  \textcolor{black}{71.57}                &  \textcolor{black}{70.58}                & \textcolor{black}{68.62}                 &  \textcolor{black}{65.28}               & \textcolor{black}{53.98}                 &   \textcolor{black}{52.72}               &     \textcolor{black}{51.31}             &  \textcolor{black}{49.60}               & \textcolor{black}{51.81}                 &  \textcolor{black}{49.48}                 & \textcolor{black}{49.61}                  & \textcolor{black}{46.96}                  \\
\textcolor{black}{FedMSplit~\citep{chen2022fedmsplit}}                        &  \textcolor{black}{74.76}                & \textcolor{black}{73.32}                  &  \textcolor{black}{72.77}                &   \textcolor{black}{71.65}              &  \textcolor{black}{55.57}                &   \textcolor{black}{53.61}               & \textcolor{black}{52.65}                 &  \textcolor{black}{51.80}               & \textcolor{black}{52.97}                  & \textcolor{black}{51.65}                  &  \textcolor{black}{50.57}                 & \textcolor{black}{48.29}                  \\ \hline
\textbf{Ours (\textbf{MFCPL})}             & \textbf{77.44} & \textbf{75.61} & \textbf{75.19} & \textcolor{black}{\textbf{73.93}}                & \textbf{56.30} & \textbf{55.69} & \textbf{55.04} &  \textcolor{black}{\textbf{53.88}}               & \textcolor{black}{\textbf{54.71}}                  & \textcolor{black}{\textbf{53.62}}                & \textcolor{black}{\textbf{52.18}}                &   \textcolor{black}{\textbf{50.26}}              \\ \hline
\end{tabular}
}%
\label{mainresult}
\end{table*}

\begin{table*}[]
\caption{{\textbf{Performance comparison of our \textbf{MFCPL} with SOTA methods} on UCI-HAR, Hateful Memes and MELD datasets under various zero-filling data rates u with $q=0.5$. The best results are marked in \textbf{bold}.}}
\centering
\resizebox{18cm}{!}{%
\begin{tabular}{c|ccc|ccc|ccc}
\hline
\multirow{2}{*}{\textbf{Methods}} & \multicolumn{3}{c|}{UCI-HAR (F1 score)}          & \multicolumn{3}{c|}{Hateful Memes (AUC score)}   & \multicolumn{3}{c}{MELD (UAR score)}             \\
                                  & $u = 0.2$        & $u = 0.5$        & $u = 0.7$        & $u = 0.2$        & $u = 0.5$        & $u = 0.7$        & $u = 0.2$        & $u = 0.5$        & $u = 0.7$        \\ \hline
FedAvg~\citep{mcmahan2017communication}                            & 74.71          & 73.55          & 72.64          & 56.34          & 55.32          & 54.77          &  \textcolor{black}{53.63}               &\textcolor{black}{52.91}                 & \textcolor{black}{52.37}                \\
FedProx~\citep{li2020federated}                            & 74.34          & 73.35          & 71.97          & 57.67          & 56.78          & 55.47          &  \textcolor{black}{53.82}               &  \textcolor{black}{53.18}               & \textcolor{black}{52.88}                \\
FedOpt~\citep{reddi2020adaptive}                             & 78.45          & 77.08          & 75.86          & 57.52          & 56.91          & 55.21          & \textcolor{black}{53.90}                & \textcolor{black}{52.69}                 & \textcolor{black}{51.79}                \\
MOON~\citep{li2021model}                              & 73.45          & 72.57          & 73.27          & 57.03          & 56.52          & 54.50          & \textcolor{black}{53.55}                & \textcolor{black}{53.02}                & \textcolor{black}{52.86}                \\ \hline
FedProto~\citep{tan2022fedproto}                          & 76.33          & 76.02          & 75.83          & 54.98          & 53.81          & 54.23          &  \textcolor{black}{54.19}               & \textcolor{black}{53.40}                &  \textcolor{black}{52.68}               \\
FedProc~\citep{mu2023fedproc}                      & 74.91          & 72.80          & 72.36          & 54.17          & 53.29          & 53.11          &  \textcolor{black}{53.06}               & \textcolor{black}{52.82}                   &  \textcolor{black}{52.04}               \\
FPL~\citep{huang2023rethinking}                             & 76.92          & 76.17          & 75.23          & 55.59          & 55.26          & 54.61          &  \textcolor{black}{54.38}               & \textcolor{black}{53.45}                        & \textcolor{black}{52.83}                        \\ \hline
\textcolor{black}{FedIoT~\citep{zhao2022multimodal}}                              & \textcolor{black}{75.82}                 & \textcolor{black}{74.13}                 &\textcolor{black}{72.03}                  & \textcolor{black}{56.27}                 & \textcolor{black}{55.16}                 & \textcolor{black}{54.19}                 &   \textcolor{black}{53.87}              &   \textcolor{black}{52.64}              &    \textcolor{black}{51.89}             \\
\textcolor{black}{FedMSplit~\citep{chen2022fedmsplit}}                           & \textcolor{black}{76.46}                  & \textcolor{black}{76.03}                  & \textcolor{black}{74.98}                &  \textcolor{black}{57.32}               &  \textcolor{black}{56.44}                & \textcolor{black}{55.78}                 & \textcolor{black}{54.05}                &  \textcolor{black}{53.27}                &     \textcolor{black}{53.02}             \\ \hline
\textbf{Ours (MFCPL)}             & \textbf{78.84} & \textbf{77.87} & \textbf{78.03} & \textbf{58.35} & \textbf{57.49} & \textbf{56.83} & \textcolor{black}{\textbf{55.36}}                &  \textcolor{black}{\textbf{55.06}}                     & \textcolor{black}{\textbf{54.81}}                 \\ \hline
\end{tabular}
}
\label{additionalresult}
\end{table*}
\subsubsection{Model Architecture} In the case of modalities such as image, text, video and audio, we follow~\citep{feng2023fedmultimodal} to extract features for local training. This method involves deriving features instead of training from scratch using raw data, a choice made to alleviate the computational burden. \textcolor{black}{Specifically, for the visual modality such as image and video, MobileNetV2~\citep{howard2017mobilenets} is employed to extract representations, while textual features are derived using MobileBERT~\citep{sun2020mobilebert}. For audio data, Mel-frequency cepstral coefficients (MFCCs)~\citep{chen2023exploring} are utilized to extract relevant features. For other modalities, raw data is utilized for training process. }
Our model architecture is composed of three key modules: the encoder, the fusion layer, and the classifier. For audio, accelerometer, and gyroscope data, the encoder adopts a 1D Conv+RNN architecture. The 1D Conv includes three CNN layers followed by two MLP layers, while the RNN component incorporates a single GRU layer. \textcolor{black}{In contrast, the encoder for images is built with two MLP layers, and for text and video, we employ an RNN architecture with a single GRU layer.} The fusion layer is implemented as a self-attention fusion layer, and the classifier is subsequently followed by two MLP layers. The hidden size for GRU layer is set as $128$ for all datasets. Regarding the projection head,  we utilize a single MLP layer with the default dimension set to $64$. We provide the detailed model architecture for each dataset in Table~\ref{ucihar_model},~\ref{hateful_model},~\ref{meld_model}.

\subsubsection{Implementation Details} 
Following the methodology outlined in~\citep{feng2023fedmultimodal}, we set a uniform missing rate of $q=\{0.5,0.7,0.8,1.0\}$ for each modality based on Bernoulli distribution, creating a scenario that simulates the presence of missing modalities. \textcolor{black}{The Bernoulli distribution with parameter $1-q$ determines the presence, and $q$ determines the absence of each modality for each client. For example, if $q=0.7$, each modality has a $70\%$ probability of being missing and $30\%$ probability of being retained. For each modality $m$, if a sample from the Bernoulli distribution is $0$, the modality is retained; if it is $1$, the modality is considered missing. In the case where $q=1.0$, each client will possess only data from one modality while other modalities are missing (cross-modal setting). This approach ensures a uniform missing rate while providing the randomized missing patterns to mimic real-world scenarios.} \textcolor{black}{Table~\ref{tab:client_distribution} illustrates the client distribution across three datasets with different missing rates. For instance, in MELD dataset ($M=3$), there are possible $7$ types of clients: audio, text, video, audio-text, audio-video, text-video, and audio-text-video.} Furthermore, to simulate the scenario where clients can access a specified proportion of the missing modality's data, we randomly sample the amount of data to be missing with the zero-filling data rate $u$. Note that, the missing modality will be filled by zeros as the input. For instance, in the case with a modality missing rate $q=0.5$, clients can access the missing modality's data with zero-filling data rate $u$ (e.g, $u=0.2$ indicates $20\%$ data in missing modality is absent). The communication round is set to $200$, and the local training epoch is $1$ for all datasets. The default value for the temperature $\tau$ is $0.1$. The batch size is set to $16$, and we employ the SGD~\citep{robbins1951stochastic} optimizer with a weight decay of $1e-5$, maintaining a consistent learning rate of $0.05$ across all datasets. 
We set $\alpha_{reg}=1$, $\alpha_{con}=2$, and $\alpha_{align}=0.1$. For the Hateful Memes dataset, we use the AUC score as the evaluation metric. The UAR score serves as the metric for the MELD dataset, and the F1 score is employed as the metric for the UCI-HAR dataset. We repeat the experiment $5$ times and report mean values.
\subsubsection{Baselines} For evaluation, we compare our \textbf{MFCPL} with several state-of-the-art FL methods in MFL setting: (1) \textbf{FedAvg}~\citep{mcmahan2017communication}, (2) \textbf{FedProx}~\citep{li2020federated}, (3) \textbf{FedOpt}~\citep{reddi2020adaptive}, (4) \textbf{MOON}~\citep{li2021model}, prototype-based FL methods: (5) \textbf{FedProto}~\citep{tan2022fedproto} (with parameter averaging), (6) \textbf{FedProc}~\citep{mu2023fedproc}, (7) \textbf{FPL}~\citep{huang2023rethinking}, \textcolor{black}{and multimodal federated learning methods: (8) \textbf{FedIoT}~\citep{zhao2022multimodal}, (9) \textbf{FedMSplit}~\citep{chen2022fedmsplit}.}


\begin{figure*}[t]
    \centering
    \begin{subfigure}{0.33\textwidth}
        \centering
        \includegraphics[width=\linewidth]{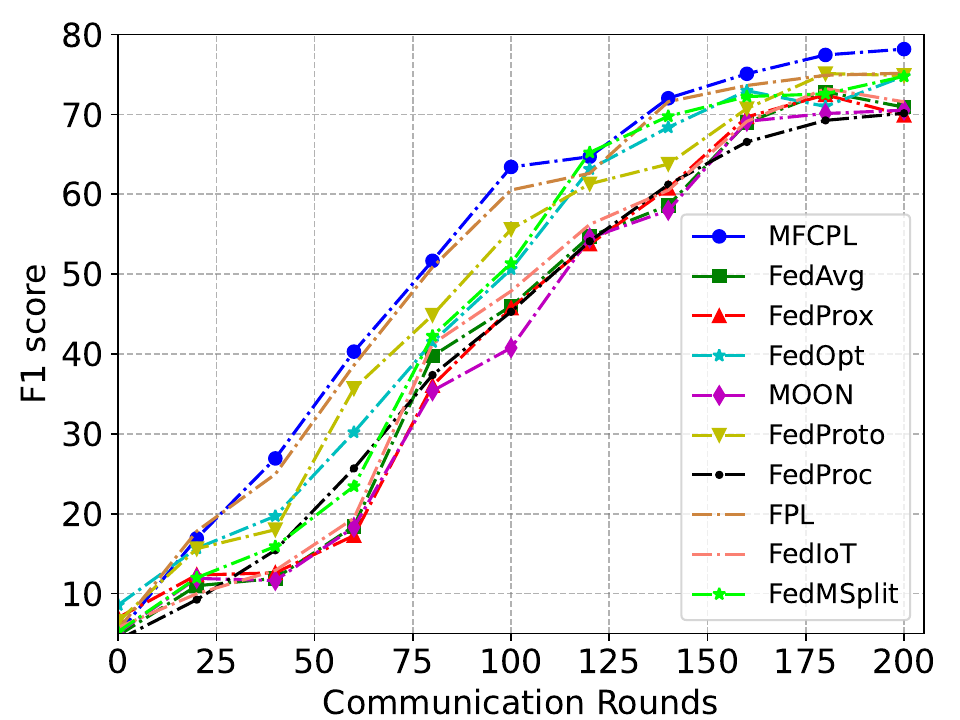}
        \caption{UCI-HAR}
        \label{fig:temperature_ucihar}
    \end{subfigure}%
    \hfill
    \begin{subfigure}{0.33\textwidth}
        \centering
        \includegraphics[width=\linewidth]{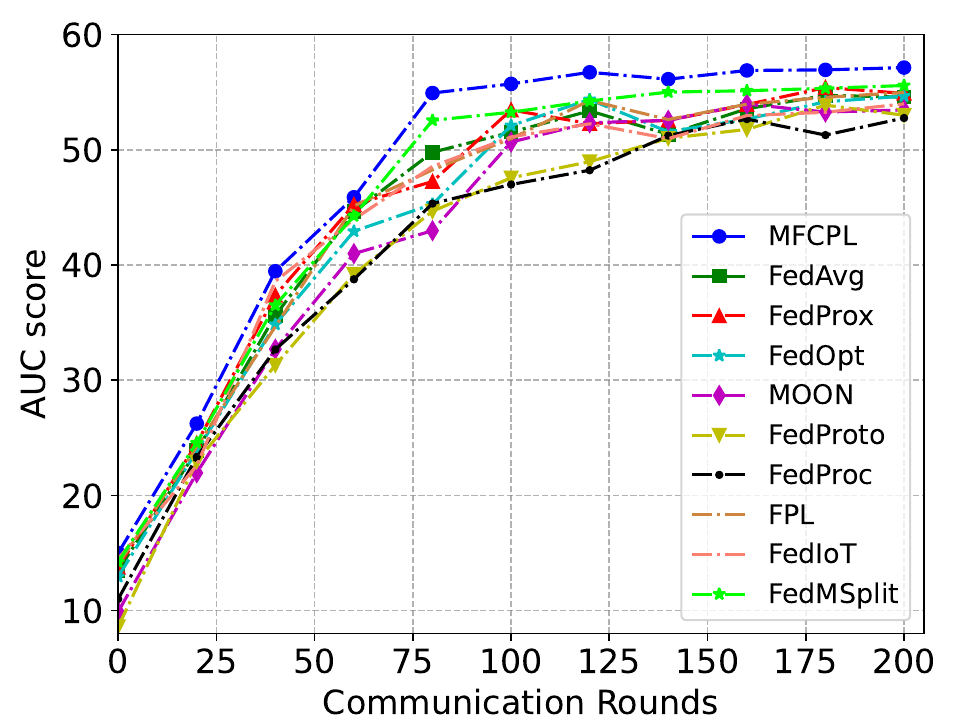}
        \caption{Hateful Memes}
        \label{fig:dimension_hatefulmemes}
    \end{subfigure}%
    \hfill
    \begin{subfigure}{0.33\textwidth}
        \centering
        \includegraphics[width=\linewidth]{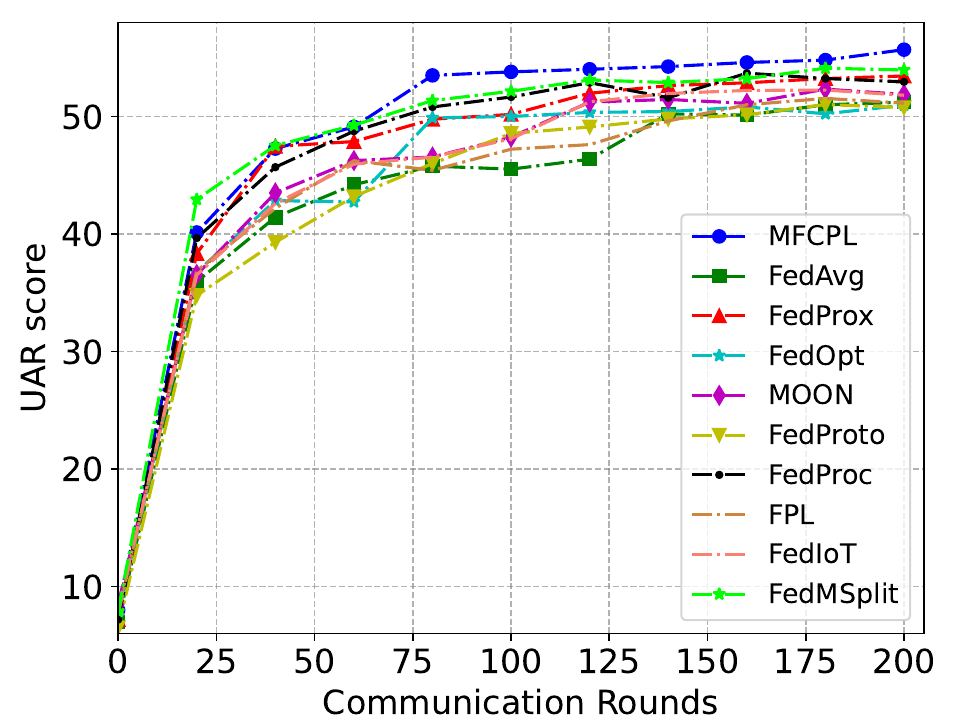}
        \caption{MELD}
        \label{fig:dimension_meld}
    \end{subfigure}
    \caption{\textcolor{black}{{Illustrative example of performance versus communication rounds three datasets with $q=0.5$.}}}
    \label{curve}
\end{figure*}

\begin{table*}[]
\caption{\textbf{Ablation study on key components of our \textbf{MFCPL}} for three datasets under various modality missing rates.}
\resizebox{18cm}{!}{%
\centering
\begin{tabular}{c|c|c|cccc|cccc|cccc}
\hline
\multirow{2}{*}{CMPR} & \multirow{2}{*}{CMPC} & \multirow{2}{*}{CMA} & \multicolumn{4}{c|}{UCI-HAR (F1 score)}                     & \multicolumn{4}{c|}{Hateful Memes (AUC score)}               & \multicolumn{4}{c}{MELD (UAR score)}                         \\
                      &                       &                      & $q = 0.5$        & $q = 0.7$        & $q = 0.8$        & $q = 1.0$        & $q = 0.5$        & $q = 0.7$        & $q = 0.8$        & $q = 1.0$        & $q = 0.5$        & $q = 0.7$        & $q = 0.8$        & $q = 1.0$        \\ \hline
\xmark                   & \xmark                   & \xmark                   & 70.90          & 69.46          & 67.50          &   \textcolor{black}{66.85}               & 54.56          & 50.09          & 49.56          &  \textcolor{black}{47.66}                & \textcolor{black}{51.22}           & \textcolor{black}{50.62}           & \textcolor{black}{50.36}           & \textcolor{black}{47.46}                 \\
\xmark                   & \cmark                  & \cmark                 & 75.09          & 74.31          & 73.63          & \textcolor{black}{72.44}               & 53.96          & 54.92          & 54.15          & \textcolor{black}{52.39}               & \textcolor{black}{52.65}          & \textcolor{black}{51.98}          & \textcolor{black}{50.91}          & \textcolor{black}{49.53}                \\
\cmark                  & \xmark                    & \cmark                  & 71.05          & 70.55          & 71.68          & \textcolor{black}{70.95}                 & 51.42          & 51.58          & 50.23          &  \textcolor{black}{49.98}               & \textcolor{black}{51.94}          & \textcolor{black}{50.82}          & \textcolor{black}{50.70}          & \textcolor{black}{48.88}                \\
\cmark                  & \cmark                  & \xmark                  & 73.56          & 70.96          & 70.41          &  \textcolor{black}{70.02}               & 51.79          & 51.17          & 50.63          &  \textcolor{black}{50.16}                  & \textcolor{black}{51.98}         & \textcolor{black}{51.15}          & \textcolor{black}{50.91}          & \textcolor{black}{49.82}                \\ \hline
\cmark                   & \cmark                    & \cmark                  & \textbf{77.44} & \textbf{75.61} & \textbf{75.19} & \textcolor{black}{\textbf{73.93}}                & \textbf{56.30} & \textbf{55.69} & \textbf{55.04} & \textcolor{black}{\textbf{53.88}}                & \textcolor{black}{\textbf{54.71}} & \textcolor{black}{\textbf{53.62}} & \textcolor{black}{\textbf{52.18}} & \textcolor{black}{\textbf{50.26}}                \\ \hline
\end{tabular}
}
\label{ablation}
\end{table*}
\subsection{Performance Comparison} \label{performance_comparison}
Table~\ref{mainresult} illustrates the performance comparison of our proposed MFCPL with other state-of-the-art methods on three multimodal federated datasets. Our method, MFCPL, consistently outperforms other baselines across various missing rates, highlighting its effectiveness in achieving better generalized global models and supporting clients with missing data. As the missing rate parameter $q$ increases, MFCPL maintains its performance, while other baselines exhibit a significant decrease in accuracy. Taking the UCI-HAR dataset with $q=0.5$ as an illustrative example, our method surpasses the best-performing baseline FPL by a notable margin, showcasing an improvement of $2.28\%$. Similar to the UCI-HAR dataset, MFCPL not only outperforms other methods on different datasets but also maintains its performance as the modality missing rate increases. Notably, when $80\%$ data of modality is missing, our method achieves improvement up to $5.48\%$ and $5.24\%$ in performance on Hateful Memes and MELD datasets, respectively. \textcolor{black}{Furthermore, under a $100\%$ missing rate, where clients possess data from only a single modality, our method demonstrates improvements up to $18.1\%$, $6,73\%$ and $13.46\%$ in performance on UCI-HAR, Hateful Memes and MELD datasets, respectively.} These results underscore the robustness and effectiveness of MFCPL by leveraging the complete prototypes to guide the local training in handling missing data scenarios across different client types and datasets. Moreover, MFCPL depicts the key advantages of maintaining the high accuracy even the modality missing rate increase, demonstrating the strong adaptability to scenarios with severe modality absences. 
Figure~\ref{curve} depicts the performance curves (performance versus communication rounds) of our MFCPL and SOTA baselines on three datasets, from which we can observe that MFCPL demonstrates a more stable convergence compared to other methods, particularly under non-IID and missing modality conditions. 

\textcolor{black}{Furthermore, we explore an alternative practical setting for MFL under missing modalities, wherein clients have access to a portion of the data in the missing modality. Table~\ref{additionalresult} presents a performance comparison of our proposed method with other state-of-the-art approaches across various data rates $u=\{0.2,0.5,0.7\}$. Similar to the scenario with $100\%$ missing data rate, MFCPL consistently outperforms other baselines across different zero-filling data rates, highlighting its effectiveness in diverse practical settings. Specifically, our proposed scheme achieves the improvements ranging from $1.5\%$ to $4\%$ compared to other baselines across most scenarios. Overall, as the zero-filling data rate increases, all methods across all datasets exhibit a decrease in performance due to the prevalence of zero-filled data. However, MFCPL maintains comparable performance even as the data rate rises, demonstrating the importance of integrating components at both modality-shared and modality-specific levels.   }



\begin{table}[]
\centering
\caption{\textbf{Performance analysis of MFCPL} on utilizing complete prototypes and unimodal prototypes.}
\resizebox{9cm}{!}{%
\begin{tabular}{c|cccc}
\hline
\multirow{2}{*}{Prototype} & \multicolumn{4}{c}{UCI-HAR}                      \\
                           & $q = 0.5$        & $q = 0.7$        & $q = 0.8$        & $q = 1.0$        \\ \hline
Unimodal Prototypes        & 77.09          & 74.67          & 73.34          & \textcolor{black}{71.65}                 \\
Complete Prototypes        & \textbf{77.44} & \textbf{75.61} & \textbf{75.19} & \textcolor{black}{\textbf{73.93}}                 \\ \hline
\multirow{2}{*}{Prototype} & \multicolumn{4}{c}{Hateful Memes}                \\
                           & $q = 0.5$        & $q = 0.7$        & $q = 0.8$        & $q = 1.0$        \\ \hline
Unimodal Prototypes        & 54.95          & 54.16          & 53.68          &   \textcolor{black}{{52.31}}                \\
Complete Prototypes        & \textbf{56.30} & \textbf{55.69} & \textbf{55.04} &   \textcolor{black}{\textbf{53.88}}                \\ \hline
\multirow{2}{*}{Prototype} & \multicolumn{4}{c}{MELD}                         \\
                           & $q = 0.5$        & $q = 0.7$        & $q = 0.8$        & $q = 1.0$        \\ \hline
Unimodal Prototypes        & \textcolor{black}{53.32}          & \textcolor{black}{52.43}            & \textcolor{black}{50.41}            & \textcolor{black}{49.14}                   \\
Complete Prototypes        & \textcolor{black}{\textbf{54.71}} & \textcolor{black}{\textbf{53.62}} & \textcolor{black}{\textbf{52.18}} & \textcolor{black}{\textbf{50.26}}                 \\ \hline
\end{tabular}
}
\label{completeproto}
\end{table}


\subsection{Comparison to Unimodal Prototypes}
\subsubsection{Details on MFCPL with Unimodal Prototypes}
\textcolor{black}{We utilize the projection head $g_1$ to map the unimodal representations $z_m$ from encoders $f_m$ into shared representation space $\mathbb{R}^d$:
\begin{align}
c_{m}=g_1(z_m)\in\mathbb{R}^d, \forall m\in \{1,2,...,M\}. 
\end{align}
We define the $k^{th}$ class prototypes for each modality as:
\begin{align}
    u^k_{m_{i}}=\frac{1}{|S^k_{i}|}\sum_{i\in S^k_{i}}c_{m_{i}}, \forall m\in \{1,2,...,M\},
\end{align}
where $S^k_i$ denotes the subset of $D_i$ belonging to $k^{th}$ . The unimodal prototypes on the server for each modality are formulated as follows:
\begin{align}
\begin{split}
    U_{m}^k&=\frac{1}{N}\sum_{i\in N}u^k_{m_{i}}\in\mathbb{R}^d,\forall m\in \{1,2,...,M\}    \\
      U_{m}&=[ U_{m}^1, U_{m}^2,\dots, U_{m}^K]    \label{aggproto_A},
\end{split}
\end{align}
%
where $U_{m}^k$ represents the unimodal prototype of each modality $m$ belonging to class $k\in K$. The calibration is conducted between the client type and its corresponding modality's prototype following the MFCPL framework.}

\begin{figure*}[]
    \centering
    \begin{subfigure}{0.33\textwidth}
        \centering
        \includegraphics[width=\linewidth]{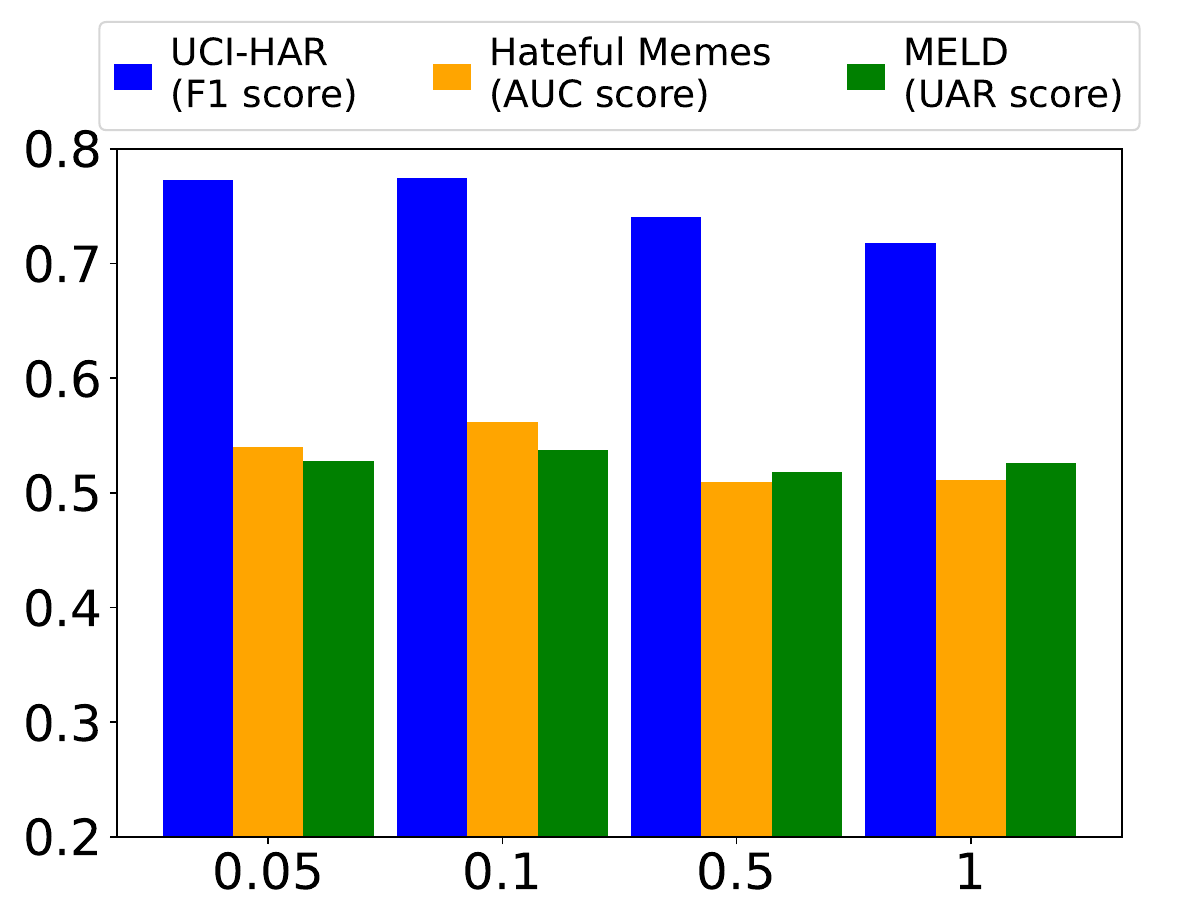}
        \caption{$q=0.5$}
        \label{fig:temperature_ucihar}
    \end{subfigure}%
     \hspace{0.02\textwidth}
    \begin{subfigure}{0.33\textwidth}
        \centering
        \includegraphics[width=\linewidth]{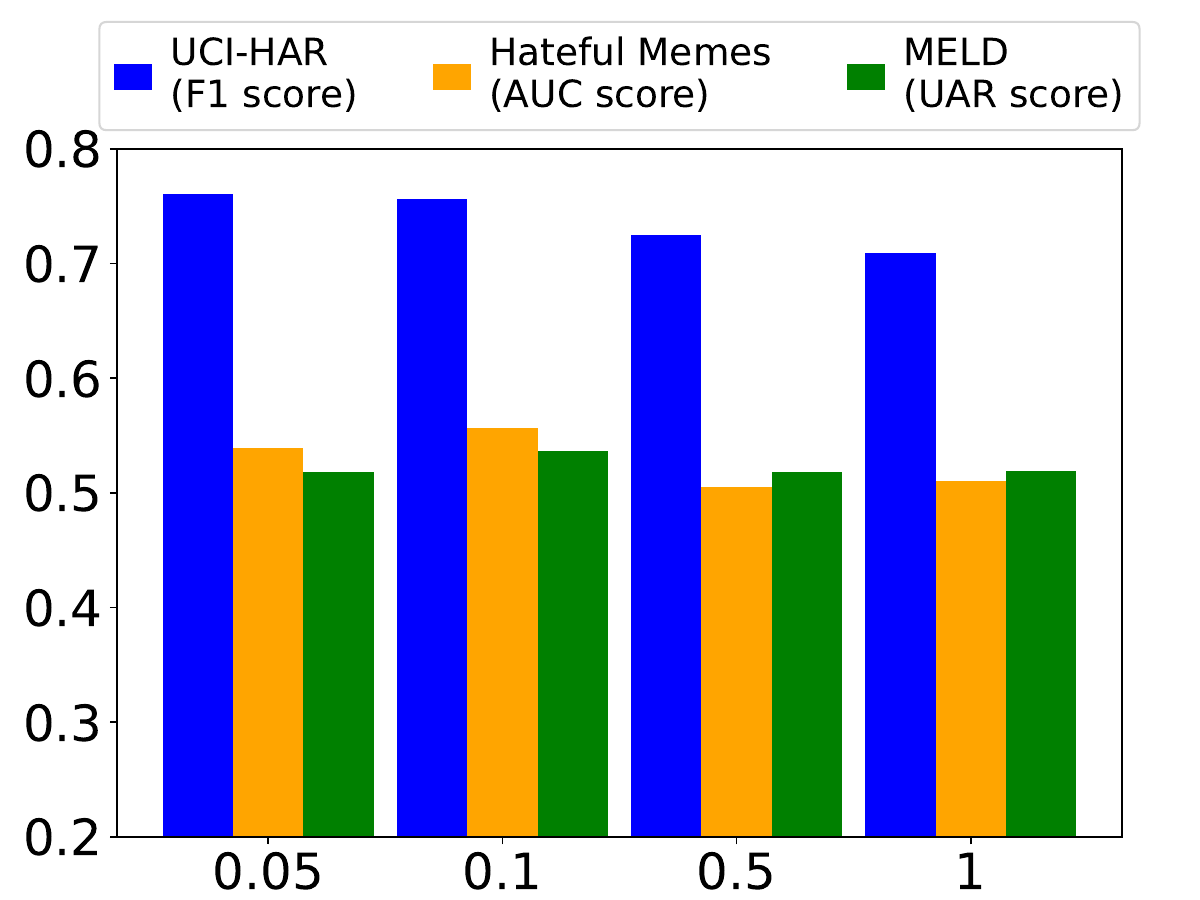}
        \caption{$q=0.7$}
        \label{fig:dimension_hatefulmemes}
    \end{subfigure}%
    \vskip\baselineskip
    \begin{subfigure}{0.33\textwidth}
        \centering
        \includegraphics[width=\linewidth]{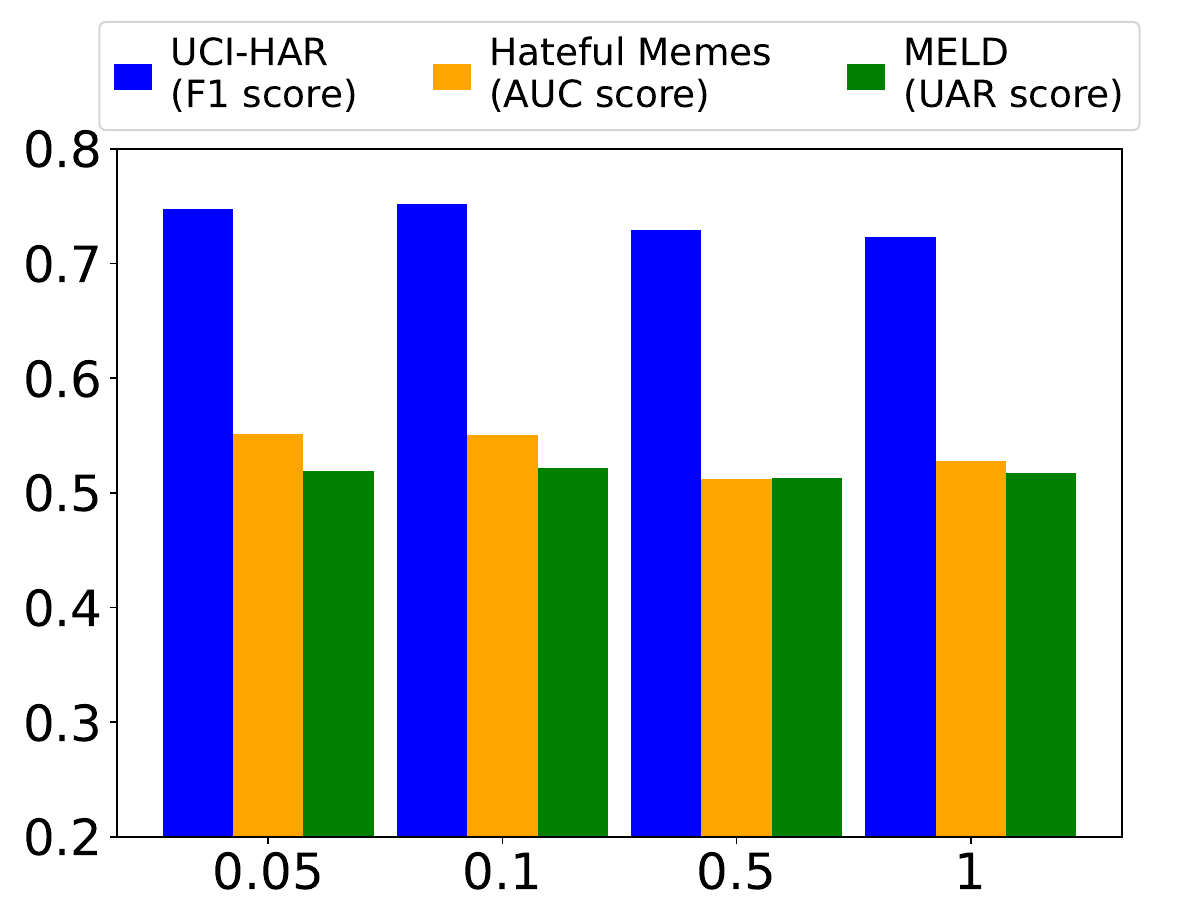}
        \caption{$q=0.8$}
        \label{fig:dimension_meld}
    \end{subfigure}%
     \hspace{0.02\textwidth}
    \begin{subfigure}{0.33\textwidth}
        \centering
        \includegraphics[width=\linewidth]{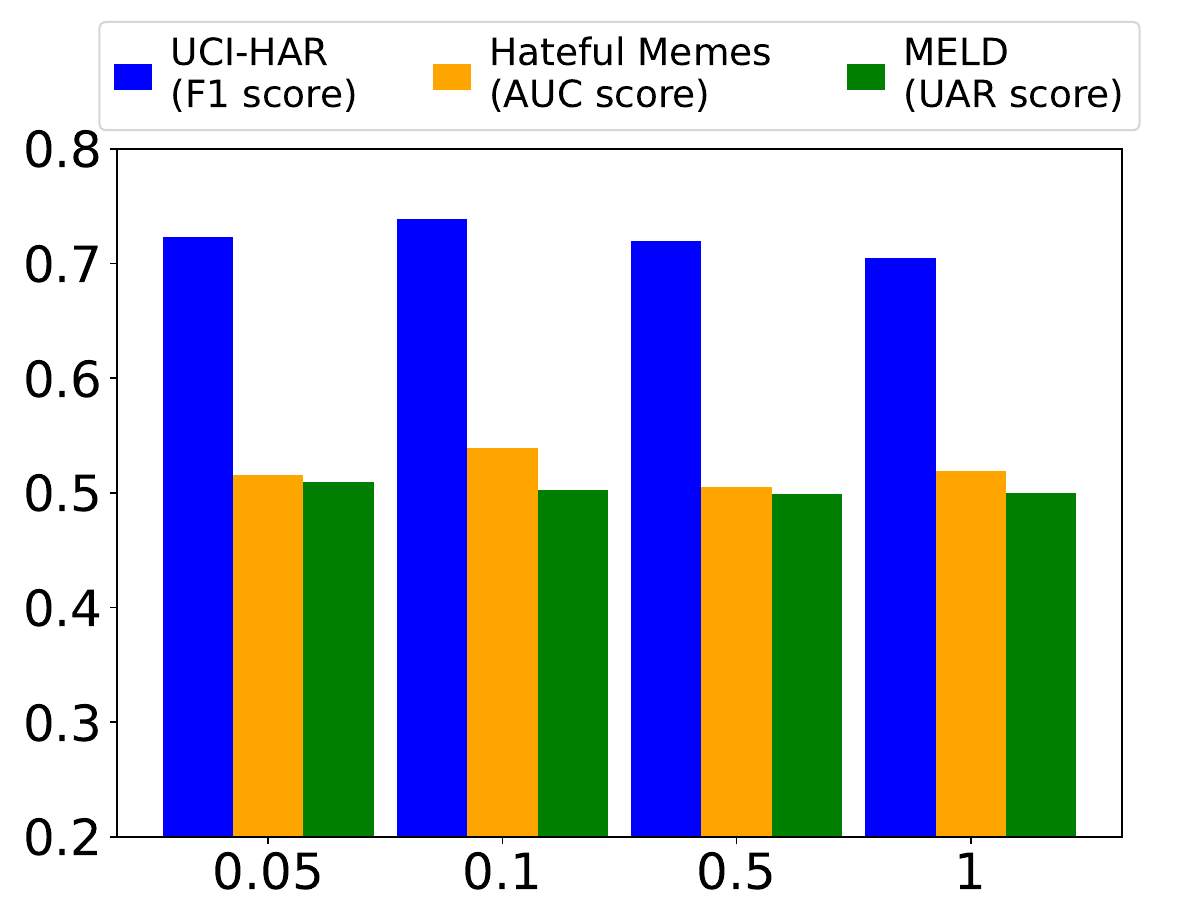}
        \caption{$q=1.0$}
        \label{fig:dimension_q1}
    \end{subfigure}%
    \caption{\textcolor{black}{\textbf{Performance analysis of MFCPL} on three datasets under different values of temperature $\tau$ with various missing rates.}}
    \label{fig:temperature}
\end{figure*}

\begin{figure*}[]
    \centering
    \begin{subfigure}{0.33\textwidth}
        \centering
        \includegraphics[width=\linewidth]{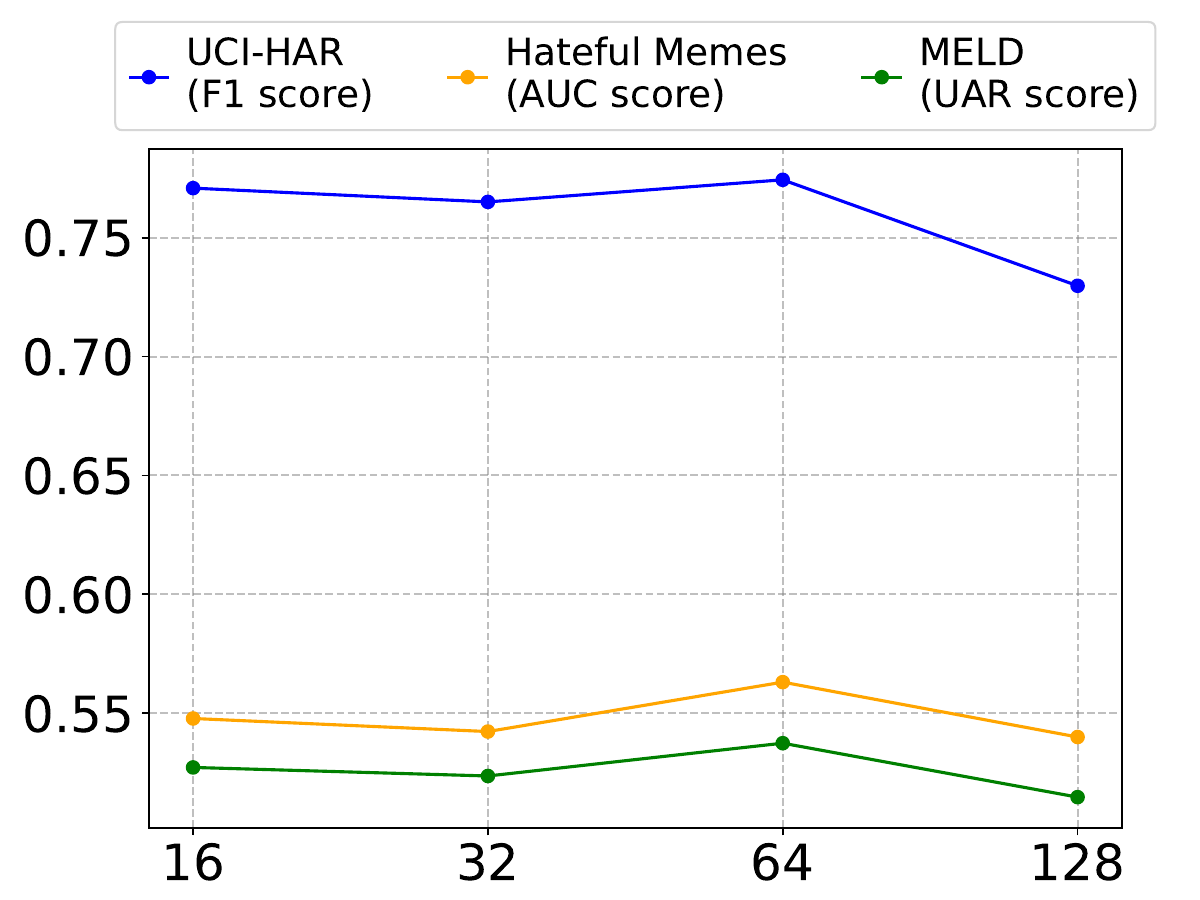}
        \caption{$q=0.5$}
        \label{fig:temperature_ucihar}
    \end{subfigure}%
    \hspace{0.02\textwidth} 
    \begin{subfigure}{0.33\textwidth}
        \centering
        \includegraphics[width=\linewidth]{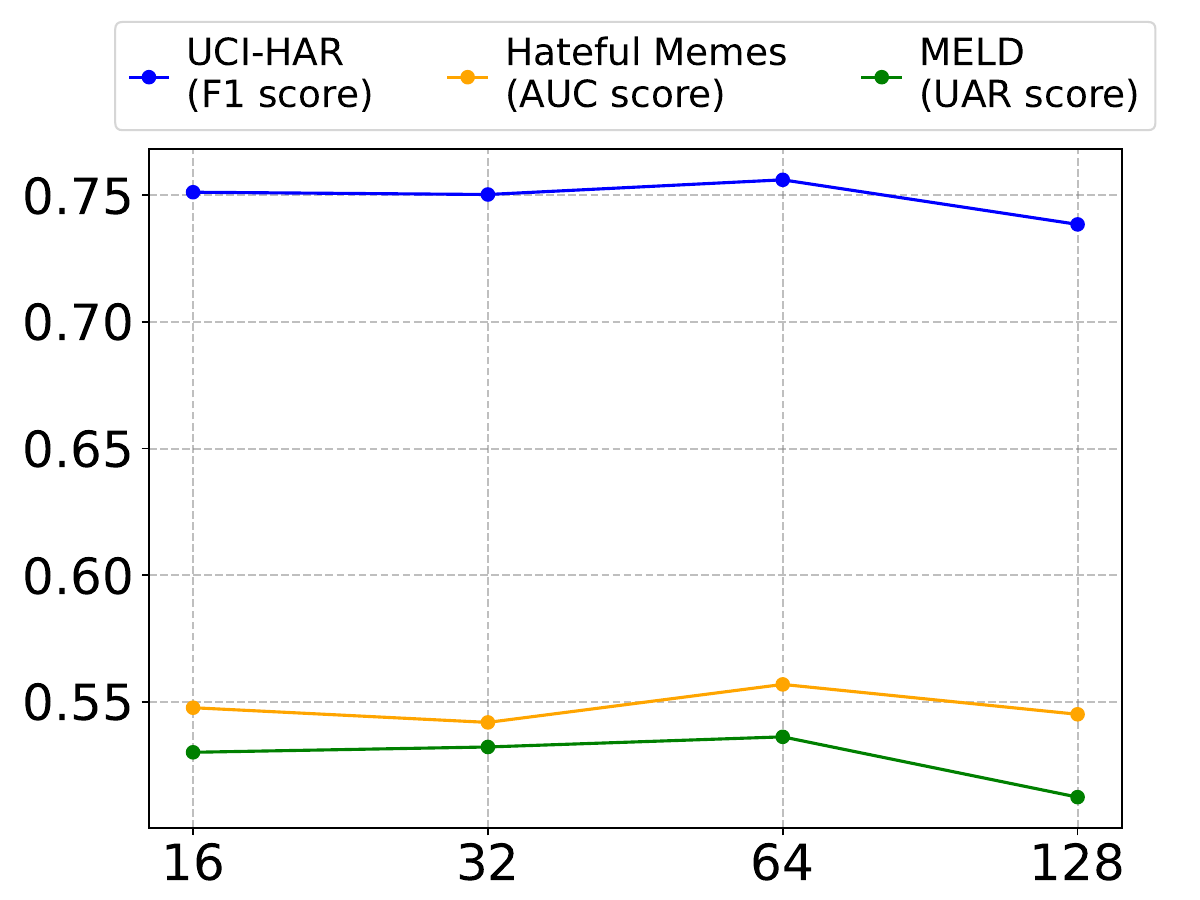}
        \caption{$q=0.7$}
        \label{fig:dimension_hatefulmemes}
    \end{subfigure}%
    \vskip\baselineskip
    \begin{subfigure}{0.33\textwidth}
        \centering
        \includegraphics[width=\linewidth]{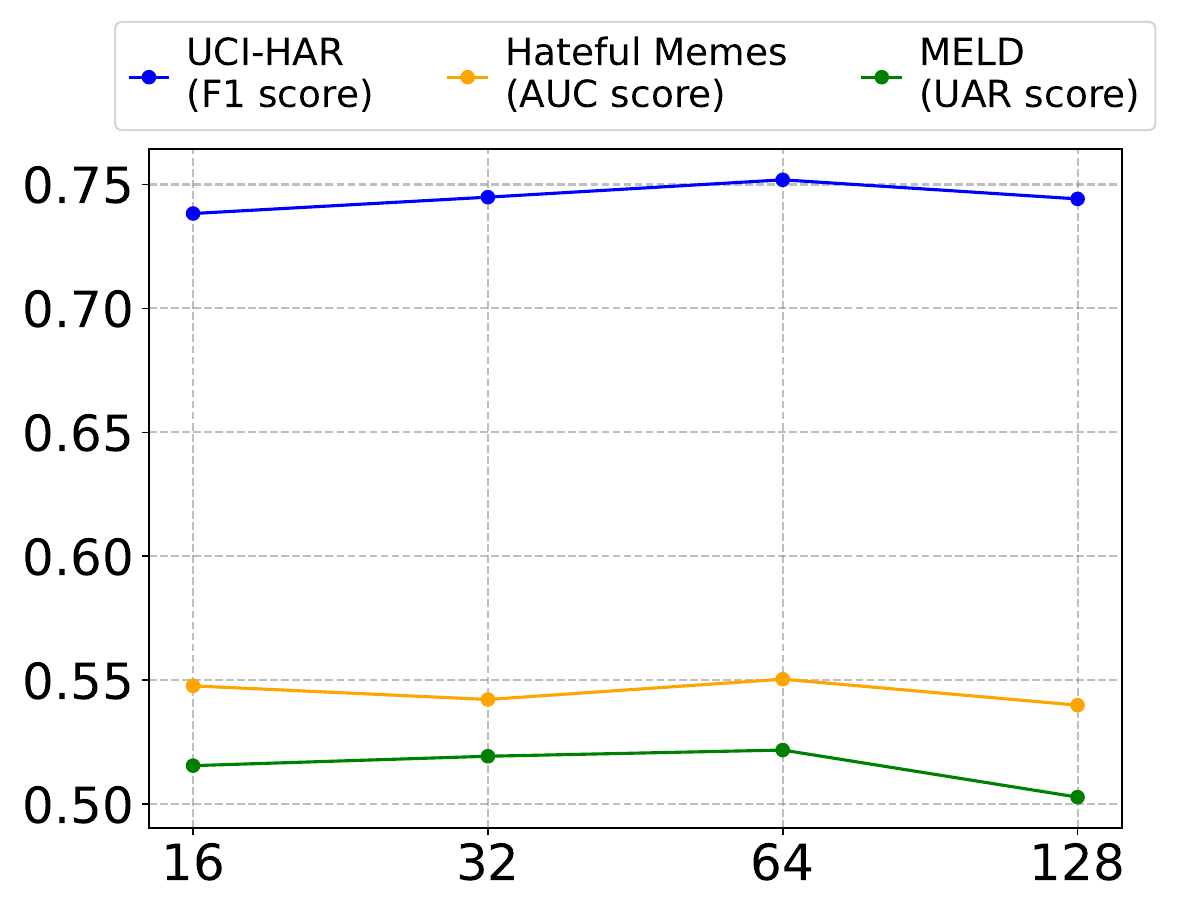}
        \caption{$q=0.8$}
        \label{fig:dimension_meld}
    \end{subfigure}%
    \hspace{0.02\textwidth} 
    \begin{subfigure}{0.33\textwidth}
        \centering
        \includegraphics[width=\linewidth]{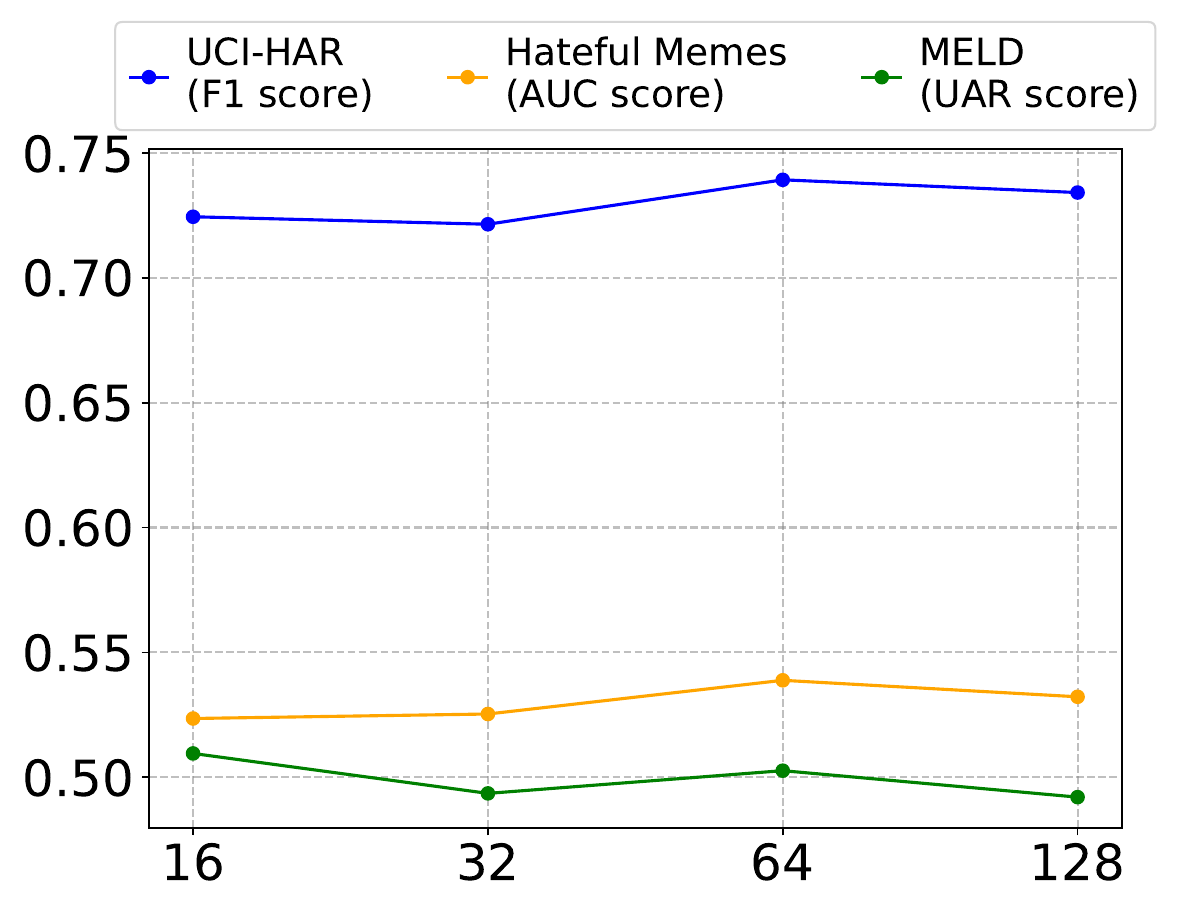}
        \caption{$q=1.0$}
        \label{fig:dimension_q1}
    \end{subfigure}%
    \caption{\textcolor{black}{\textbf{Performance analysis of MFCPL} on three datasets under different projection head dimensions $d$ with various missing rates.}}
    \label{fig:dimension}
\end{figure*}
\subsubsection{Performance Comparison}
To assess the efficacy of our proposed complete prototypes, we conducted experiments comparing them with unimodal prototypes on various missing rates, as summarized in Table~\ref{completeproto}. The results demonstrate the superior performance of utilizing our proposed complete prototypes compared to the unimodal prototypes. Notably, our method exhibits a performance improvement of $1.85\%$, $1.36\%$ and $1.77\%$ on UCI-HAR, Hateful Memes and MELD dataset over the unimodal prototypes with $80\%$ data of modality is missing, respectively. This finding highlights the effectiveness of capturing diverse modality knowledge from complete prototypes, providing valuable guidance for the training process. 
\subsection{\textcolor{black}{Computation and Communication Scalability}}
\textcolor{black}{To evaluate the computational cost of our method, we provide a detailed comparison of the computational costs of each additional component (i.e., CMPR, CMPC, and CMA) with the baseline computation cost of FedAvg. Computational complexity is measured in floating-point operations per second (FLOPs) using the FlopCountAnalysis() function from the fvcore library of Meta\footnote{https://github.com/facebookresearch/fvcore}. Table~\ref{tab:cost_computation} illustrates the computational cost of each component on the MELD dataset, which includes three modalities (audio, text, and video).} 
\textcolor{black}{As shown in Table~\ref{tab:cost_computation}, the baseline FedAvg incurs a computational cost of 1,421M FLOPs, whereas the combined cost of CMPR, CMPC, and CMA components for prototype computation and alignment is negligible by comparison—approximately 453 times smaller than the CE component. This demonstrates that our proposed design introduces minimal additional computational cost relative to the baseline algorithm (e.g., FedAvg).}

\textcolor{black}{Regarding scalability, the computation cost at each client remains constant regardless of the number of federated clients. On the server side, an increase in the number of clients results in more prototypes being transmitted, leading to slightly higher communication costs. However, the size of prototypes is significantly smaller than the model size, making the added communication overhead negligible. Table~7 details the communication costs for each component on the MELD dataset. The results demonstrate that the computational and communication costs of prototype computation are minimal for both the server and clients.}
\textcolor{black}{Based on these observations, our method demonstrates scalability even when applied to large-scale systems with real-world datasets or a large number of clients. Importantly, the computation cost at each client scales with the size of the local data rather than the total dataset size, ensuring that resource-constrained edge devices can efficiently handle the computation. }

\begin{table}[]
\centering
\textcolor{black}{
\caption{The computation cost of different components on MELD dataset.}
\label{tab:cost_computation}
\begin{tabular}{|c|c|c|c|c|}
\hline
\textbf{Component} & \textbf{FedAvg} & \textbf{CMPR} & \textbf{CMPC} & \textbf{CMA} \\ \hline
\textbf{\# FLOPs} &1,421M  &2.36M  &0.39M  &0.39M  \\ \hline
\end{tabular}
}
\end{table}

\begin{table}[]
\centering
\textcolor{black}{
\caption{The communication cost (MB) of different components on the MELD dataset. In this experiment, we evaluate with $8$ participating clients in $200$ communication rounds.}
\label{tab:cost_communication}
\begin{tabular}{|c|c|c|c|}
\hline
\textbf{Component} & \makecell{\textbf{Model} \\ \textbf{Parameters}} & \makecell{\textbf{Local} \\ \textbf{Prototypes}} & \makecell{\textbf{Global} \\ \textbf{Prototypes}} \\ \hline
\textbf{Cost (MB)} & $860$ & $0.36$ & $0.045$ \\ \hline
\end{tabular}
}
\end{table}
\subsection{Ablation Studies} \label{sec:ablation}
\subsubsection{Effects of Key Components}To comprehensively evaluate the impact of individual components on the performance of MFCPL, we conducted an ablation study by selectively removing each component: CMPR, CMPC, and CMA. The results are summarized in Table~\ref{ablation}. Our findings indicate that leveraging all three components leads to the optimal performance across all three datasets, each characterized by distinct missing rates.  Significantly, CMPC and CMA demonstrate a more pronounced impact compared to CMPR. This observation highlights the critical importance of ensuring alignment between complete prototypes and modality-specific representations, as well as the effectiveness of noise reduction in supporting clients with missing modalities. Additionally, we observed that the contribution of CMPR positively impacts the learning process, providing further insights into its role in enhancing overall performance.

\begin{table}[ht]
\centering
\caption{\textbf{Performance analysis of MFCPL} on cross-modal alignment  with $q=0.5$.}
\begin{tabular}{c|ccc}
\hline
Case      & UCI-HAR        & Hateful Memes  & MELD           \\ \hline
KL        & 74.86          & 54.46          & \textcolor{black}{52.63}          \\
L1        & 76.62          & 55.42          & \textcolor{black}{53.42}          \\
Smooth L1 & 74.57          & 55.18          & \textcolor{black}{\textbf{54.95}} \\
L2        & \textbf{77.44} & \textbf{56.30} & \textcolor{black}{54.71}          \\ \hline
\end{tabular}
\label{alignment_table}
\end{table}

\begin{table}[t]
\caption{Model architecture used in our experiments for UCI-HAR dataset.}
\begin{center}
\textcolor{black}{
\resizebox{8cm}{!}{%
\begin{tabular}{@{}ll|lll@{}}
\toprule
\multicolumn{2}{c|}{\multirow{2}{*}{\textbf{Components}}} & \multicolumn{3}{c}{\multirow{2}{*}{\textbf{Layer Detail}}} \\
\multicolumn{2}{c|}{}                                     & \multicolumn{3}{c}{}                                             \\ \midrule
\multicolumn{2}{l|}{\multirow{8}{*}{\makecell{Accelerometer Encoder \\ (1D Conv+RNN)}}} 
    & \multicolumn{3}{l}{Conv1d(3, 32, kernel\_size=5, stride=1, padding=2)}       \\
\multicolumn{2}{l|}{}                                     & \multicolumn{3}{l}{Conv1d(32, 64, kernel\_size=5, stride=1, padding=2)}      \\
\multicolumn{2}{l|}{}                                     & \multicolumn{3}{l}{Conv1d(64, 128, kernel\_size=5, stride=1, padding=2)}      \\
\multicolumn{2}{l|}{}                                     & \multicolumn{3}{l}{ReLU()}      \\
\multicolumn{2}{l|}{}                                     & \multicolumn{3}{l}{MaxPool1d(kernel\_size=2, stride=2)}      \\
\multicolumn{2}{l|}{}                                     & \multicolumn{3}{l}{Dropout(p=0.1)}      \\
\multicolumn{2}{l|}{}                                     & \multicolumn{3}{l}{GRU(128, 128, batch\_first=True, dropout=0.1)}      \\ \midrule
\multicolumn{2}{l|}{\multirow{8}{*}{\makecell{Gyroscope Encoder \\ (1D Conv+RNN)}}} 
    & \multicolumn{3}{l}{Conv1d(3, 32, kernel\_size=5, stride=1, padding=2)}       \\
\multicolumn{2}{l|}{}                                     & \multicolumn{3}{l}{Conv1d(32, 64, kernel\_size=5, stride=1, padding=2)}      \\
\multicolumn{2}{l|}{}                                     & \multicolumn{3}{l}{Conv1d(64, 128, kernel\_size=5, stride=1, padding=2)}      \\
\multicolumn{2}{l|}{}                                     & \multicolumn{3}{l}{ReLU()}      \\
\multicolumn{2}{l|}{}                                     & \multicolumn{3}{l}{MaxPool1d(kernel\_size=2, stride=2)}      \\
\multicolumn{2}{l|}{}                                     & \multicolumn{3}{l}{Dropout(p=0.1)}      \\
\multicolumn{2}{l|}{}                                     & \multicolumn{3}{l}{GRU(128, 128, batch\_first=True, dropout=0.1)}      \\ \midrule
\multicolumn{2}{l|}{\multirow{3}{*}{Self-Attention }}           
    & \multicolumn{3}{l}{Linear(128, 512)}       \\
\multicolumn{2}{l|}{}                                     & \multicolumn{3}{l}{Tanh()}      \\
\multicolumn{2}{l|}{}                                     & \multicolumn{3}{l}{Linear(512, 6)}      \\ \midrule
\multicolumn{2}{l|}{\multirow{4}{*}{Classifier Head}}           
    & \multicolumn{3}{l}{Linear(768, 64)}       \\
\multicolumn{2}{l|}{}                                     & \multicolumn{3}{l}{ReLU()}      \\
\multicolumn{2}{l|}{}                                     & \multicolumn{3}{l}{Dropout(p=0.1)}      \\
\multicolumn{2}{l|}{}                                     & \multicolumn{3}{l}{Linear(64, 6)}      \\ \midrule
\multicolumn{2}{l|}{\multirow{2}{*}{Projection Layers}}           
    & \multicolumn{3}{l}{Linear(768, 64) for projector}       \\
\multicolumn{2}{l|}{}                                     & \multicolumn{3}{l}{Linear(128, 64) for projector2}      \\ \bottomrule
\end{tabular} 
}}
\label{ucihar_model}
\end{center}
\end{table}

\begin{table}[!]
\caption{Model architecture used in our experiments for Hateful Memes dataset.}
\label{tab:img_text_classifier_architecture}
\begin{center}
\textcolor{black}{
\resizebox{8cm}{!}{%
\begin{tabular}{@{}ll|lll@{}}
\toprule
\multicolumn{2}{c|}{\multirow{2}{*}{\textbf{Components}}} & \multicolumn{3}{c}{\multirow{2}{*}{\textbf{Layer Detail}}} \\
\multicolumn{2}{c|}{}                                     & \multicolumn{3}{c}{}                                             \\ \midrule
\multicolumn{2}{l|}{\multirow{4}{*}{Image Encoder}} 
    & \multicolumn{3}{l}{Linear(1280, 128)}       \\
\multicolumn{2}{l|}{}                                     & \multicolumn{3}{l}{ReLU()}      \\
\multicolumn{2}{l|}{}                                     & \multicolumn{3}{l}{Dropout(p=0.1)}      \\
\multicolumn{2}{l|}{}                                     & \multicolumn{3}{l}{Linear(128, 128)}      \\ \midrule
\multicolumn{2}{l|}{\multirow{1}{*}{Text Encoder}}           
    & \multicolumn{3}{l}{GRU(512, 128, batch\_first=True, dropout=0.1)}       \\ \midrule
\multicolumn{2}{l|}{\multirow{3}{*}{Self Attention Fusion}}           
    & \multicolumn{3}{l}{Linear(128, 512)}       \\
\multicolumn{2}{l|}{}                                     & \multicolumn{3}{l}{Tanh()}      \\
\multicolumn{2}{l|}{}                                     & \multicolumn{3}{l}{Linear(512, 6)}      \\ \midrule
\multicolumn{2}{l|}{\multirow{4}{*}{Classifier Head}}           
    & \multicolumn{3}{l}{Linear(768, 64)}       \\
\multicolumn{2}{l|}{}                                     & \multicolumn{3}{l}{ReLU()}      \\
\multicolumn{2}{l|}{}                                     & \multicolumn{3}{l}{Dropout(p=0.1)}      \\
\multicolumn{2}{l|}{}                                     & \multicolumn{3}{l}{Linear(64, 2)}      \\ \midrule
\multicolumn{2}{l|}{\multirow{2}{*}{Projection Layers}}           
    & \multicolumn{3}{l}{Linear(768, 64) for projector}       \\
\multicolumn{2}{l|}{}                                     & \multicolumn{3}{l}{Linear(128, 64) for projector2}      \\ \bottomrule
\end{tabular} 
}}
\end{center}
\label{hateful_model}
\end{table}

\begin{table}[]
\caption{Model architecture used in our experiments for MELD dataset.}
\begin{center}
\textcolor{black}{
\resizebox{8cm}{!}{%
\begin{tabular}{@{}ll|lll@{}}
\toprule
\multicolumn{2}{c|}{\multirow{2}{*}{\textbf{Components}}} & \multicolumn{3}{c}{\multirow{2}{*}{\textbf{Layer Detail}}} \\
\multicolumn{2}{c|}{}                                     & \multicolumn{3}{c}{}                                             \\ \midrule
\multicolumn{2}{l|}{\multirow{7}{*}{\makecell{Audio Encoder \\ (1D Conv+RNN)}}} 
    & \multicolumn{3}{l}{Conv1d(80, 32, kernel\_size=5, stride=1, padding=2)}       \\
\multicolumn{2}{l|}{}                                     & \multicolumn{3}{l}{Conv1d(32, 64, kernel\_size=5, stride=1, padding=2)}      \\
\multicolumn{2}{l|}{}                                     & \multicolumn{3}{l}{Conv1d(64, 128, kernel\_size=5, stride=1, padding=2)}      \\
\multicolumn{2}{l|}{}                                     & \multicolumn{3}{l}{ReLU()}      \\
\multicolumn{2}{l|}{}                                     & \multicolumn{3}{l}{MaxPool1d(kernel\_size=2, stride=2)}      \\
\multicolumn{2}{l|}{}                                     & \multicolumn{3}{l}{Dropout(p=0.3)}      \\
\multicolumn{2}{l|}{}                                     & \multicolumn{3}{l}{GRU(128, 128, batch\_first=True, dropout=0.3)}      \\ \midrule
\multicolumn{2}{l|}{\multirow{1}{*}{Text Encoder}}           
    & \multicolumn{3}{l}{GRU(512, 128, batch\_first=True, dropout=0.3)}       \\ \midrule
\multicolumn{2}{l|}{\multirow{1}{*}{Video Encoder}}           
    & \multicolumn{3}{l}{GRU(1280, 128, batch\_first=True, dropout=0.3)}       \\ \midrule
\multicolumn{2}{l|}{\multirow{3}{*}{Self-Attention Fusion}}           
    & \multicolumn{3}{l}{Linear(128, 512)}       \\
\multicolumn{2}{l|}{}                                     & \multicolumn{3}{l}{Tanh()}      \\
\multicolumn{2}{l|}{}                                     & \multicolumn{3}{l}{Linear(512, 6)}      \\ \midrule
\multicolumn{2}{l|}{\multirow{4}{*}{Classifier Head}}           
    & \multicolumn{3}{l}{Linear(768, 64)}       \\
\multicolumn{2}{l|}{}                                     & \multicolumn{3}{l}{ReLU()}      \\
\multicolumn{2}{l|}{}                                     & \multicolumn{3}{l}{Dropout(p=0.3)}      \\
\multicolumn{2}{l|}{}                                     & \multicolumn{3}{l}{Linear(64, 4)}      \\ \midrule
\multicolumn{2}{l|}{\multirow{2}{*}{Projection Layers}}           
    & \multicolumn{3}{l}{Linear(768, 64) for projector}       \\
\multicolumn{2}{l|}{}                                     & \multicolumn{3}{l}{Linear(128, 64) for projector2}      \\ \bottomrule
\end{tabular} 
}}
\end{center}
\label{meld_model}
\end{table}

\subsubsection{Temperature parameter} We further explore the overall performance under different values of $\tau$ with different missing rates, as depicted in Figure~\ref{fig:temperature}. Notably, we observe that a smaller temperature tends to enhance the training process more effectively than a larger one. It is worth noting that excessively high temperatures may exhibit a slightly negative effect on the overall performance across various missing rates.

\subsubsection{Dimension of projection head} \label{subsec:dimension} We vary the dimension of the projection head $d$ and present the results in Figure~\ref{fig:dimension}. Our observations indicate that the MFCPL method exhibits robustness across different dimension sizes $d$. However, a noteworthy trend emerges: as we increase the dimension to larger sizes, there is a slight decrease in performance. This finding suggests that a smaller projection head dimension can minimize redundant information and enhance performance. \textcolor{black}{It is intriguing to note that even a small projected space dimension ($d=16$) can yield better performance than a larger dimension ($d=128$). This highlights the effectiveness of even a compact projection space in preserving correlations between different representations from various modalities. For instance, in the Hateful Memes dataset, where the original feature space for image and text is $[1280,512]$, such small projection spaces prove to be beneficial.}

\subsubsection{Ablation study on cross-modal alignment loss}
\textcolor{black}{In this experiment, we delve into the examination of cross-modal alignment loss using various types of loss functions across three datasets, as depicted in Table~\ref{alignment_table}, including Kullback-Leibler (KL) Divergence~\citep{kullback1997information}, L1, and smooth L1 loss functions. Our results illustrate that the utilizing L2 loss (MSE loss) predominantly yields superior performance across most datasets. However, it is noteworthy that for datasets like MELD or UCI-HAR, L1 and smooth L1 loss functions also achieve notable performance, such as with smooth L1 loss on the MELD dataset. This observation underscores the adaptability of our cross-modal alignment approach to different loss functions, showcasing its efficacy across varied functions and mitigate the difference of multiple modalities. }

\section{Conclusion}
In this paper, we propose \textbf{MFCPL}, a novel multimodal federated learning framework, breaking through the twin barriers of data heterogeneity and severely missing modalities. Our approach leverages a unique complete prototype methodology, incorporating three key components. We initiate the alignment of complete prototypes at two levels: modality-specific and modality-shared representations, thereby extracting rich semantic representation from multiple modalities. Additionally, we integrate cross-modal alignment to alleviate noise stemming from missing modalities, resulting in significant performance gains. Experiments show that our MFCPL consistently outperforms other state-of-the-art methods under severely missing modality scenarios, achieving up to $7.69\%$, $5.48\%$ and $5.24\%$ improvements in performance for UCI-HAR, Hateful Memes, and MELD datasets, respectively, when $80\%$ data of modality is missing. Thus, MFCPL paves the new way for robust multimodal learning in real-world settings with diverse and incomplete data.



\bibliographystyle{elsarticle-harv} 
\bibliography{reference}






\end{document}